\documentclass[conference]{IEEEtran}
\IEEEoverridecommandlockouts
\usepackage{cite}
\usepackage{amsmath,amssymb,amsfonts}
\usepackage{algorithmic}
\usepackage{graphicx}
\usepackage{textcomp}
\usepackage{xcolor}
\usepackage{float}
\usepackage{amsmath}
\usepackage{multirow}
\usepackage{algorithm, algorithmic}
\usepackage{mathtools}
\usepackage{array}
\usepackage{subcaption}
\usepackage[utf8]{inputenc}
\usepackage{xspace}
\usepackage{hyperref}
\hypersetup{colorlinks,urlcolor=blue}
\usepackage{placeins}

\def\BibTeX{{\rm B\kern-.05em{\sc i\kern-.025em b}\kern-.08em
    T\kern-.1667em\lower.7ex\hbox{E}\kern-.125emX}}
\begin{document}

\title{MixBoost: Synthetic Oversampling with Boosted Mixup for Handling Extreme Imbalance}

% \newcommand{\IEEEauthorblockN}[1]{#1}
% \newcommand{\IEEEauthorblockA}[1]{#1}

% \author{
% \IEEEauthorblockN{Anubha Kabra\thanks{Contributed during internship at MDSR}}
% \IEEEauthorblockA{\textit{Delhi Technological University}\\}\\
% \and
% \IEEEauthorblockN{Ayush Chopra\thanks{Corresponding Author}}
% \IEEEauthorblockA{\textit{Media and Data Science Research Lab, Adobe}\\}\\
% \and
% \IEEEauthorblockN{Nikaash Puri}
% \IEEEauthorblockA{\textit{Media and Data Science Research Lab, Adobe}\\}\\
% \and
% \IEEEauthorblockN{Pinkesh Badjatiya}
% \IEEEauthorblockA{\textit{Media and Data Science Research Lab, Adobe}\\}\\
% \and
% \IEEEauthorblockN{Sukriti Verma}
% \IEEEauthorblockA{\textit{Media and Data Science Research Lab, Adobe}\\}\\
% \and
% \IEEEauthorblockN{Piyush Gupta}
% \IEEEauthorblockA{\textit{Media and Data Science Research Lab, Adobe}\\}\\
% \and
% \IEEEauthorblockN{Balaji K}
% \IEEEauthorblockA{\textit{Media and Data Science Research Lab, Adobe}\\}
% }

\author{
\IEEEauthorblockN{Anubha Kabra\IEEEauthorrefmark{1}\IEEEauthorrefmark{4}\thanks{\IEEEauthorrefmark{1}Work done during internship at MDSR},
Ayush Chopra\IEEEauthorrefmark{3}\thanks{\IEEEauthorrefmark{3}Work done while at Adobe},
Nikaash Puri\IEEEauthorrefmark{2},
Pinkesh Badjatiya\IEEEauthorrefmark{2},
Sukriti Verma\IEEEauthorrefmark{2},\\
Piyush Gupta\IEEEauthorrefmark{2}
and
Balaji K\IEEEauthorrefmark{2}
},
\IEEEauthorblockA{\IEEEauthorrefmark{2}Media and Data Science Research Lab, Adobe}
\IEEEauthorblockA{\IEEEauthorrefmark{3}Massachusetts Institute of Technology}
\IEEEauthorblockA{\IEEEauthorrefmark{4}Delhi Technological University}
% \IEEEauthorblockA{\IEEEauthorrefmark{4}Tyrell Inc., 123 Replicant Street, Los Angeles, California 90210--4321}
}

% \maketitle

%% To use when \citet{} is not supported
\newcommand{\citet}[1]{\cite{#1}}

\newcommand{\OURNAME}{\textit{MixBoost}\xspace}

\newcommand{\gmean}{{\relax\ifmmode{g\text{-}mean}\else{$g$-$mean$~}\fi}} 

\newcommand{\todo}[1]{}

\newcolumntype{D}{@{}>{\lrbox0}l<{\endlrbox}}

\DeclarePairedDelimiter\ceil{\lceil}{\rceil}
\DeclarePairedDelimiter\floor{\lfloor}{\rfloor}

\maketitle

\begin{abstract}
Training a classification model on a dataset where the instances of one class outnumber those of the other class is a challenging problem. Such imbalanced datasets are standard in real-world situations such as fraud detection, medical diagnosis, and computational advertising. We propose an iterative data augmentation method, \OURNAME, which intelligently selects ($Boost$) and then combines ($Mix$) instances from the majority and minority classes to generate synthetic hybrid instances that have characteristics of both classes.
We evaluate \OURNAME on 20 benchmark datasets, show that it outperforms existing approaches and test its efficacy through significance testing. We also present ablation studies to analyze the impact of the different components of MixBoost. 
%Our approach is being deployed in a leading Digital Marketing suite of products. 
% In the present work, we tackle the problem of binary classification on highly imbalanced dataset. To this end we propose, ISWI, a majority-minority interpolation based strategy for generation of synthetic training instances. ISWI generates synthetic training instances by probabilistic interpolation of real instances sampled from the majority and minority classes. We also introduce an entropy-weighted technique to sample the training instances. We extensively evaluate the proposed approach on 20 benchmark imbalanced datasets against existing state-of-the-art techniques and highlight the superiority of ISWI. We also present ablation studies to highlight the impact of each of our contributions and to validate the scalability of ISWI to extreme imbalance.
\end{abstract}


\maketitle

\section{Introduction}
\label{sec:introduction}
Several real-world situations (fault detection~\cite{wang2014resampling}, disease classification~\cite{miri2018extracting}, software failures~\cite{bennin2017mahakil}, oil spill detection~\cite{kubat1997addressing} and protein sequence detection~\cite{al2005feature}) involve learning from datasets where the instances of one class (the majority class) far outnumber those of the other class (the minority class).
In certain \textbf{extremely imbalanced} datasets, the ratio of the number of instances from the two classes is very low (\textit{$<1:100 $}) or the number of instances of the minority class is very small~\cite{swim}. 
For instance, in gamma-ray anomaly detection ~\cite{sharma2018learning}, the datasets for learning have 10 anomalies and approximately 25,000 benign signatures.
In fraud detection, there are typically about 5 fraudulent examples in a dataset of over 300,000 transactions~\cite{wei2013effective}.

Training a binary classification model on such extremely imbalanced datasets is a challenging problem. A popular class of methods alleviates this problem by augmenting the training data with synthetic instances before training the classification model~\cite{chawla2002smote,chawla2003smoteboost,borderlinesmote,swim}. 
These data augmentation methods generate synthetic minority class instances using either minority or majority class instances from the training dataset. 
However, existing methods often generate instances that don't improve (or even worsen) the classifier performance.
\begin{figure}[t]
    \centering
    \includegraphics[width=0.47\textwidth]{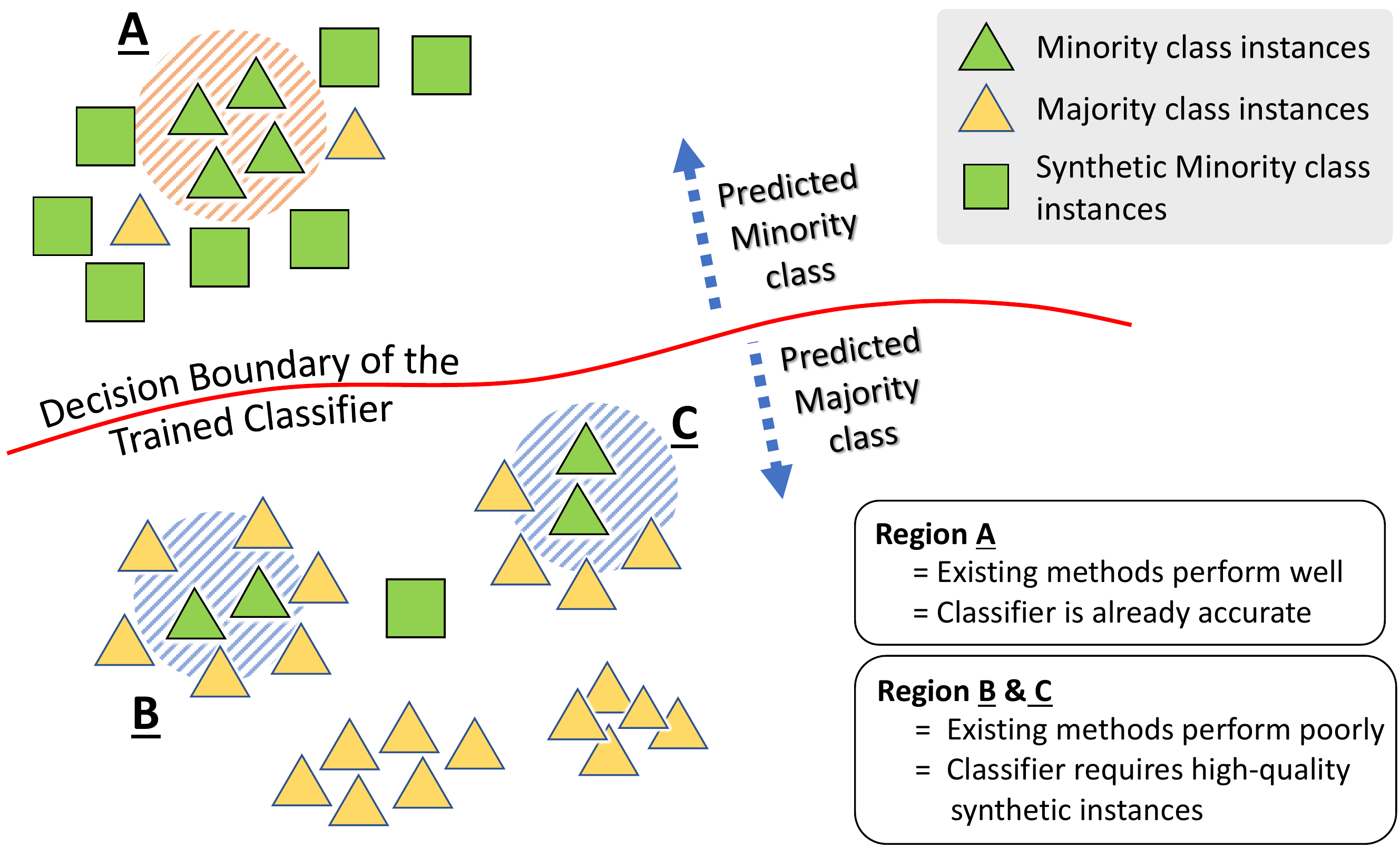}
    \caption{Illustration to describe the limitation of synthetic oversampling techniques. Existing methods (SMOTE~\cite{chawla2002smote} and its variants, SWIM~\cite{swim}) select candidate instances from one of the classes and create synthetic instances based on these selected instances. Therefore, most generated synthetic instances lie near clusters (often within the convex hull) of instances that are often already correctly classified by the classification model (\underline{region A}). Further, these methods generate fewer and poorer quality synthetic instances in regions of the input space where the model does not perform well (\underline{regions B and C}).}
    \label{fig:intro:limitation-existing-approaches-intuition}
\end{figure}%
\begin{figure*}
    \includegraphics[width=0.835\textwidth]{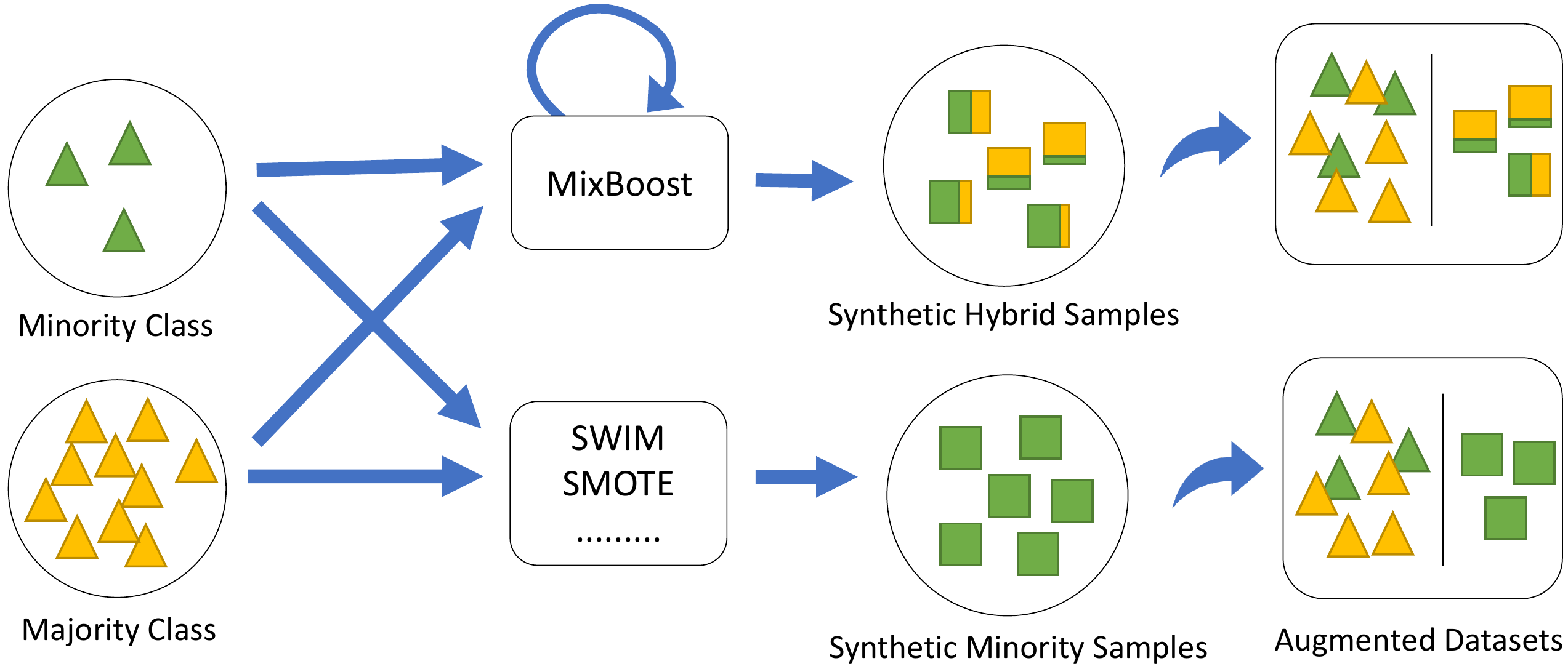}
    \centering
    \caption{
    High-level summary of the data-augmentation pipeline. 
    The triangle represents the data instances in the non-augmented training dataset. 
    Green is for minority class and yellow for majority class.
    The square represents the synthetic generated instances that are added to the training dataset (augmentation) prior to training.
    Traditional approaches such as SWIM~\cite{swim}, SMOTE~\cite{chawla2002smote}, and its variants generate synthetic minority class instances (denoted by green). 
    In contrast, \OURNAME generates synthetic hybrid instances that interpolate the minority and majority class instances (denoted by the green-yellow color scheme).
    }
    \label{figure:mixboost-high-level}
\end{figure*}
Intuitively, these methods generate good quality synthetic instances in regions of the input space where the classification model is already accurate. 
In regions where the model is inaccurate and requires synthetic instances, these methods often do not perform as well.
Figure~\ref{fig:intro:limitation-existing-approaches-intuition} illustrates this intuition. 
We believe that using the trained model (whose performance we are trying to improve) to guide the data augmentation process will allow us to augment regions of the input space where the model performs poorly. 
Further, existing methods generate synthetic \textbf{homogeneous} instances, i.e., instances that belong to a single class (usually the minority class).
Recent work in Computer Vision~\cite{mixup, bcplus, summers2019improved} has demonstrated the value of augmenting training datasets with non-homogeneous \textbf{hybrid} instances to learn more robust representations. 

We describe a data augmentation method (\OURNAME) that improves classifier performance on such imbalanced datasets. 
Our key contributions are:
\begin{itemize}
    \item We introduce a data augmentation method, \OURNAME that generates synthetic \textbf{hybrid} instances to augment an imbalanced dataset (Section \ref{sec:proposed-approach}). 
    \OURNAME has two innovative components. 
    \textbf{First}, it mixes instances of the minority and majority classes to generate synthetic hybrid instances that have elements of both classes (Section \ref{sec:mix-step}).
    \textbf{Second}, it intelligently selects the instances for mixing using a novel entropy-weighted \textbf{low-high} technique (Section~\ref{instance-sampling}).
    \item We show that \OURNAME outperforms state-of-the-art data augmentation methods on multiple highly imbalanced benchmark datasets for different levels of class imbalance (Section~\ref{results}).
\end{itemize}
There are several categories of methods to augment an imbalanced dataset prior to training a classification model. 
Under-sampling methods discard instances of the majority class at random to balance the class distribution. 
However, removing instances of the majority class can lead to a loss of information and subsequently degrade classifier performance. 
Over-sampling methods duplicate instances of the minority class at random to balance the class distribution. 
These methods add duplicates of existing instances to the training dataset and therefore do not add new information \cite{galar2011review, garcia2009evolutionary}. 
More sophisticated data augmentation methods augment the training datasets by creating synthetic minority class instances. 
For instance, SMOTE~\cite{chawla2002smote} creates synthetic minority instances by interpolating minority class instances in the training data. 
However, SMOTE ignores majority class data when creating synthetic instances. 
Variants of SMOTE (Borderline SMOTE~\cite{borderlinesmote}, ADASYN~\cite{adasyn}, SMOTEBoost~\cite{chawla2003smoteboost}) use the distribution of majority class data to filter instances created using minority class data. 
However, since these methods use only minority class data, the synthetic instances they create are restricted to the convex-hull of the minority class distribution. 
To expand the space spanned by generated instances, SWIM~\cite{swim} creates instances by inflating minority class data along the density contours of majority class data.

\newcommand*\shapeTriangle{ \protect \includegraphics[scale=0.08]{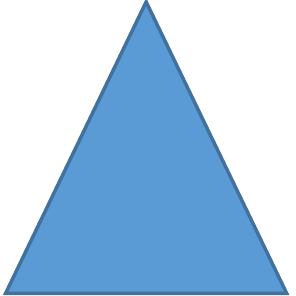}}
\newcommand*\shapeSquare{ \protect \includegraphics[scale=0.08]{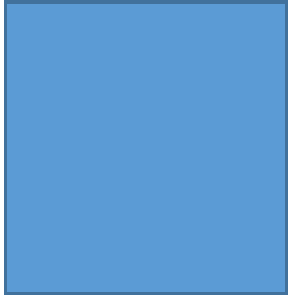}}

\begin{figure*}
    \includegraphics[width=0.95\textwidth]{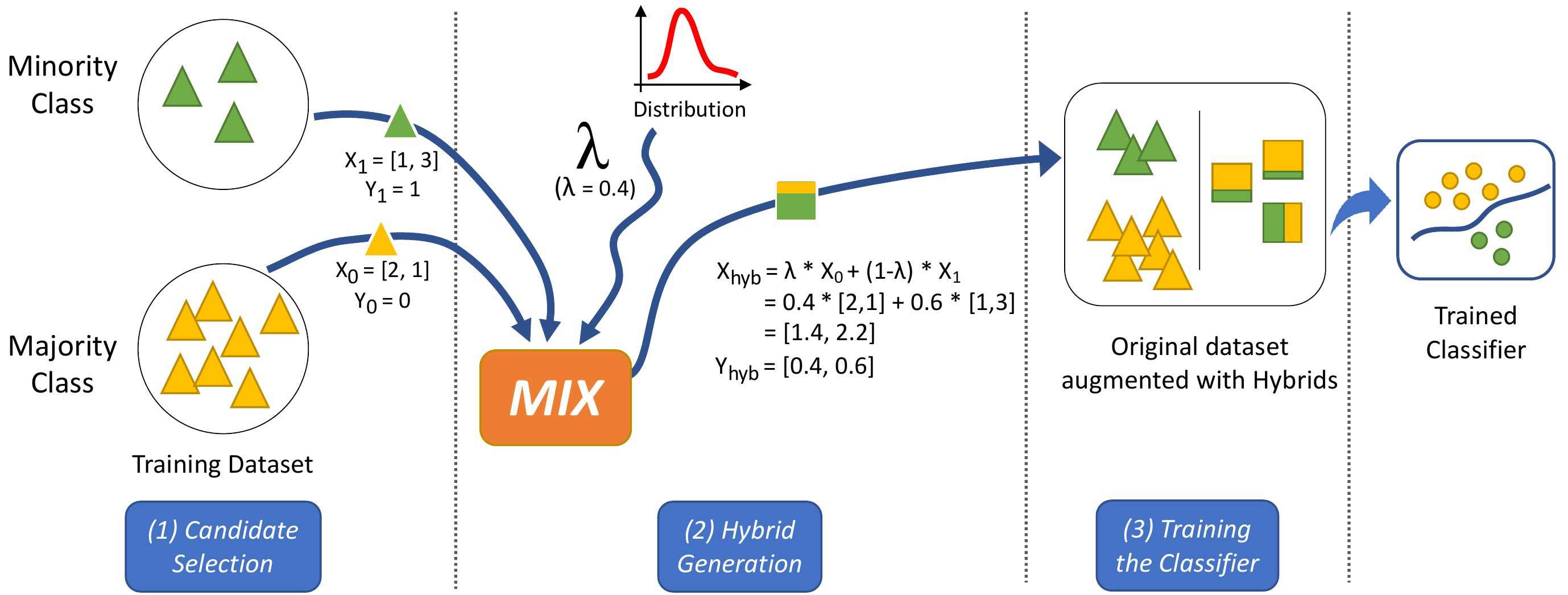}
    \centering
    \caption{
    High-level architecture of \OURNAME.
    In the (1) candidate selection step ($Boost$), \OURNAME uses entropy to intelligently select instances from the training dataset.
    In the (2) hybrid generation step ($Mix$), we mix the selected instances to create synthetic hybrid instances.
    $\lambda$ is the interpolation ratio that controls the extent of overlap between the instances to be mixed.  
    We then augment the training dataset with the generated hybrid instances and train a classifier on the augmented dataset.  
    }
    \label{figure:figure-overview}
\end{figure*}
Existing data augmentation methods give promising results in a variety of situations.
However, these methods focus on either the majority class or the minority class. 
Approaches that focus on the minority class often overlook the majority class and can degrade classifier performance by generating borderline or overlapping instances. 
On the other hand, approaches that focus on the majority class tend to overfit the training distribution. In contrast, \OURNAME combines instances from the majority and minority classes to create synthetic hybrid instances. Each synthetic hybrid instance contains elements of a majority class and a minority class instance. 
Further, \OURNAME uses an information-theoretic measure along with the trained classifier to select the instances to combine.   
Figure~\ref{figure:mixboost-high-level} illustrates the essential idea and distinction to related approaches.

Figure~\ref{figure:figure-overview} shows the high-level architecture of \OURNAME. \OURNAME generates synthetic hybrid instances iteratively. Each iteration consists of a $Boost$ step followed by a $Mix$ step. In the $Boost$ step, \OURNAME uses the trained classifier to intelligently select pairs of instances from the minority and majority class for mixing.
Intuitively, \OURNAME selects pairs of instances such that one instance in each pair is close to the decision boundary of the trained classifier, and the other instance is far from the decision boundary. In the $Mix$ step, \OURNAME mixes the selected instances to create synthetic hybrid instances that are used to augment the training dataset. The classifier is re-trained with the original data augmented with the hybrid instances before the next iteration of \OURNAME.

In Section~\ref{sec:proposed-approach} we delineate the proposed approach. Next, in Section~\ref{sec:experimental-settings} we present our experimental setup and evaluation strategy. In Section ~\ref{results} we compare quantitative performance of \OURNAME with multiple strong baselines on 20 benchmark datasets as well as test the efficacy of our framework through significance tests. In Section~\ref{sec:abalation-studies} we perform ablation studies to analyze the impact of different design choices for the proposed approach. Finally, we summarize related work in Section~\ref{sec:related-work}

\section{Approach}
\label{sec:proposed-approach}

In this section, we explain in detail the various steps of our proposed \OURNAME algorithm.
In Section~\ref{sec:approach:definitions} we define the terminologies used throughout the paper. In Section~\ref{sec:approach:high-level-summary} we provide a high-level overview of the proposed approach followed by detailed explanation of the different stages in Sections~\ref{sec:approach:instance-selection} and~\ref{sec:approach:hybrid-generation}.

\subsection{Definitions}
\label{sec:approach:definitions}
A binary classification task is characterized by a dataset $X^{n{\times}m} \in R$ and corresponding labels $Y^{n} \in \{ 0, 1 \} $. 
The class label $0$ corresponds to the majority class, and label $1$ corresponds to the minority class throughout the paper. 
Let $n$ be the number of instances in the dataset, and $m$ be the number of features for an instance in the dataset.
Then, the training objective is to learn a function, $f(x^i) \rightarrow y^i$, that maps the instances $x^i \in X$ to their corresponding labels $y^i \in Y$. 
For imbalanced datasets, the number of instances of the majority class far outnumbers that of the minority class.
The dataset, denoted by $ D^{orig}:=(X,Y)$, is split into a training dataset $ D^{orig}_{train}:=(X_{train},Y_{train})$, and a testing dataset $ D^{orig}_{test}:=(X_{test},Y_{test})$. Individual datapoints from these datasets are referred to as  $d^{orig}_{train}$ and $d^{orig}_{test}$ respectively.
Let $n_{train}$ be the number of instances in the training dataset.
We train a binary classification model $M$ on $D^{orig}_{train}$ and evaluate on the test dataset $D^{orig}_{test}$ using \gmean~\cite{kubat1997addressing} and ROC-AUC~\cite{roc_auc_ref} metrics.

\OURNAME uses the classifier $M$ along with dataset $D^{orig}_{train}$ to create synthetic hybrid instances. 
The instances are added to the training dataset to create the augmented dataset $D^{aug}_{train}$ which is then used to in-turn train the classifier $M$. This represents a single iteration of \OURNAME. \OURNAME is run for $k$ such iterations to expand the augmented dataset. We report performance by evaluating the model trained using the final augmented dataset, obtained at the end of $k$ iterations, using $D^{orig}_{test}$.

\subsection{MixBoost: High-Level Summary}
\label{sec:approach:high-level-summary}
\OURNAME generates synthetic hybrid instances by interpolating instances from the majority and minority classes.
A single iteration of \OURNAME (summarized in Figure~\ref{figure:figure-overview}) consists of the following steps:
\begin{enumerate}
    \item \textbf{Candidate Selection ($Boost$):}
    We sample candidate instances from the majority and minority classes in $D_{train}^{orig}$ prior to mixing. 
    We experiment with both random and guided sampling techniques (Section \ref{instance-sampling}). 
    \item \textbf{Hybrid Generation ($Mix$):}
    We mix the sampled instances to generate synthetic hybrid instances (Section \ref{sec:mix-step}). 
    $D^{hyb}:=(X^{hyb},Y^{hyb})$ denotes the set of synthetic instances generated in this step (and each synthetic instance is correspondingly referred to as $d^{hyb}$). 
\end{enumerate}

At the end of a \OURNAME iteration, we add $D^{hyb}$ to $D^{aug}_{train}$ and retrain the classifier $M$ on $D^{aug}_{train}$. 

For ease of explanation, we focus on generating a single synthetic hybrid instance. However, the steps to generate several instances follow from this single instance case. 
\OURNAME creates a synthetic hybrid instance by mixing a majority class instance ($x_0$) and a minority class instance ($x_1$). 
There are \textit{two} important considerations to this process.
First, how should \OURNAME select the \textit{best} candidates $x_0$ and $x_1$ from $D_{train}$? Second, how should \OURNAME \textit{intelligently} combine $x_0$ and $x_1$ to create the synthetic hybrid instance $d^{hyb}$? 
We first consider the problem of instance selection.

\subsection{MixBoost: Candidate Selection (the $Boost$ step)}\label{instance-sampling}
\label{sec:approach:instance-selection}
Unlike existing works, \OURNAME creates synthetic datapoints that are conditioned on both the majority and minority class instances. For generating these synthetic hybrids, we propose two alternative strategies for selection of the candidate instances,

\subsubsection{\textbf{R-Selection}}
First, we propose \textit{R-Selection} which randomly selects (majority and minority) candidates $x_0$ and $x_1$ from $D^{orig}_{train}$ using uniform random variables.
% For generating a synethic hybrid instance, we select $x_0$ and $x_1$ from $D^{orig}_{train}$ using Uniform Random Selection or \textit{R-Selection}. 
The probability of selecting an instance is equal for all instances within the majority and minority classes. 
Formally, if $p^{i}_{0}$ is the probability of selecting the $i^{th}$ majority class instance and $p^{j}_{1}$ is the probability of selecting the $j^{th}$ minority class instance, then:
\begin{equation}
    p^{i}_{0} = 1/n_0 \thickspace \forall i%
    , \quad \quad%
    p^{j}_{1} = 1/n_1 \thickspace \forall j%
\end{equation}
where $n_0$ is the number of majority class instances and $n_1$ is the number of minority class instances in $D_{train}^{orig}$.

\subsubsection{\textbf{Entropy Weighted (EW) Selection}}
\textit{R-Selection} assumes that all synthetic hybrid instances generated by mixing any combination of minority/majority class instances are equally helpful additions to the augmented dataset for training. However, it is possible that mixing certain combinations of minority/majority class instances from $D_{train}^{orig}$ can result in better training performance. 

For a given classification model $M$, entropy is a measure of uncertainty in the models prediction. Consequently, we introduce an alternate strategy called Entropy-Weighted selection (\textit{EW-Selection}). \textit{EW-Selection} posits that selecting candidates from $D_{train}^{orig}$ by actively weighting them with the uncertainty in their model predictions will result in generation of more useful hybrid instances.  Formally, if $\Vec{y}^{~i}_{pred}$ is the probability distribution over target classes output by the classifier $M$ for the $i^{th}$ instance in $D_{train}^{orig}$, then the entropy of the sample, denoted by $E^{i}$, is computed as,
\begin{equation}
    E^{i} = Entropy(\Vec{y}^{~i}_{pred}) = -\sum {y^{i}_{pred}*log(y^{i}_{pred})}
\end{equation}
where the summation is over the elements of the vector $\Vec{y}^{~i}_{pred}$.

For a given data instance, a high $E^{i}$ implies that model $M$ is less certain about the ground-truth class prediction of the instance while a low $E^{i}$ implies that $M$ is more certain about the ground-truth class prediction of the instance. We hypothesise that augmenting training dataset with synthetic instances in vicinity of high entropy feature sub-spaces (where model is currently uncertain about predictions) can improve model training performance.

Let $\hat{E_{0}} = \sum{E^{i}} \thickspace \forall x^i_0$ ($x^i$ with class $c_0$) is the sum of entropy values for majority class instances and $\hat{E_{1}} = \sum{E^{i}} \thickspace \forall x^i_1$ ($x^i$ with class $c_1$) denote the sum of entropy values for minority class instances.
Then, $P(x^i|c_0)$ and $P(x^j|c_1)$ denote the \textit{entropy ratios}, which are the probability of selecting a majority class instance and a minority class instance respectively for mixing. Both $P(x^i|c_0)$ and $P(x^j|c_1)$ are computed as,
\begin{equation}
    \label{eq:p_i}
    P(x^i|c_0) = \frac{E^{i}}{\hat{E_{0}}}, \quad \quad  P(x^j|c_1) = \frac{E^{j}}{\hat{E_{1}}}
\end{equation}
In practice, we find that selecting one candidate instance with \textit{high entropy ratio} and the other with \textit{low entropy ratio} leads to generation of synthetic hybrid instances that result in the best performance. \OURNAME interleaves this \textbf{low-high} selection with the generation of hybrid instances using the $Mix$ step described below.

\subsection{MixBoost: Hybrid Generation (the $Mix$ step)} \label{sec:mix-step}
\label{sec:approach:hybrid-generation}
Let ($x_0$, $y_0$) and ($x_1$, $y_1$) be instances selected (in the $Boost$ step) from the majority and minority class respectively.
We adopt the mixing strategy introduced in \cite{mixup}.
\begin{equation}
    \begin{split}
        x_{hyb} =& ~\lambda*x_0 + (1 - \lambda)*x_1 \\
        y_{hyb} =& ~\lambda*y_0 + (1 - \lambda)*y_1 
    \end{split}
\end{equation}
($x_{hyb}$, $y_{hyb}$) is the generated synthetic hybrid instance. 
$\lambda$ is the interpolation ratio.
Intuitively, $\lambda$ controls the extent of overlap between the two instances used to generate the synthetic hybrid instance. 
MIXUP~\cite{mixup} and BC+~\cite{bcplus} are existing image augmentation techniques which use linear and non-linear interpolations respectively.
BC+ samples $\lambda$ using the Uniform Distribution $U(0, 1)$ while MIXUP samples $\lambda$ using the $Beta(\alpha, \alpha)$ distribution. 
For most tasks, MIXUP uses $\alpha=1$, which also reduces to sampling $\lambda$ from a Uniform Distribution~\cite{summers2019improved}. 
In contrast, \OURNAME works with skewed data distributions arising from extreme class imbalance.
Correspondingly, we explore sampling $\lambda$ from several probability density functions and discuss our findings in Section~\ref{sec:discussion-lambda-sampling}.

The resultant synthetic hybrid instances ($D^{hyb}$) are used to augment the training dataset ($D^{orig}_{train}$), generating $D^{aug}_{train}$, which is used to (re)train the classifier. The classifier is (re)trained on a \textit{cross-entropy} loss by using the soft labels for the $D^{hyb}$ and the one-hot labels for $D^{orig}_{train}$ as ground truths. In the subsequent iterations, this classifier is used to update entropy values for candidates in the original training dataset to facilitate the next iteration of \OURNAME.  The entire workflow of our approach is shown in Figure~\ref{figure:figure-overview}.

\section{Experimental Settings}
\label{sec:experimental-settings}
\begin{figure}
    \centering
    \includegraphics[width=0.48\textwidth]{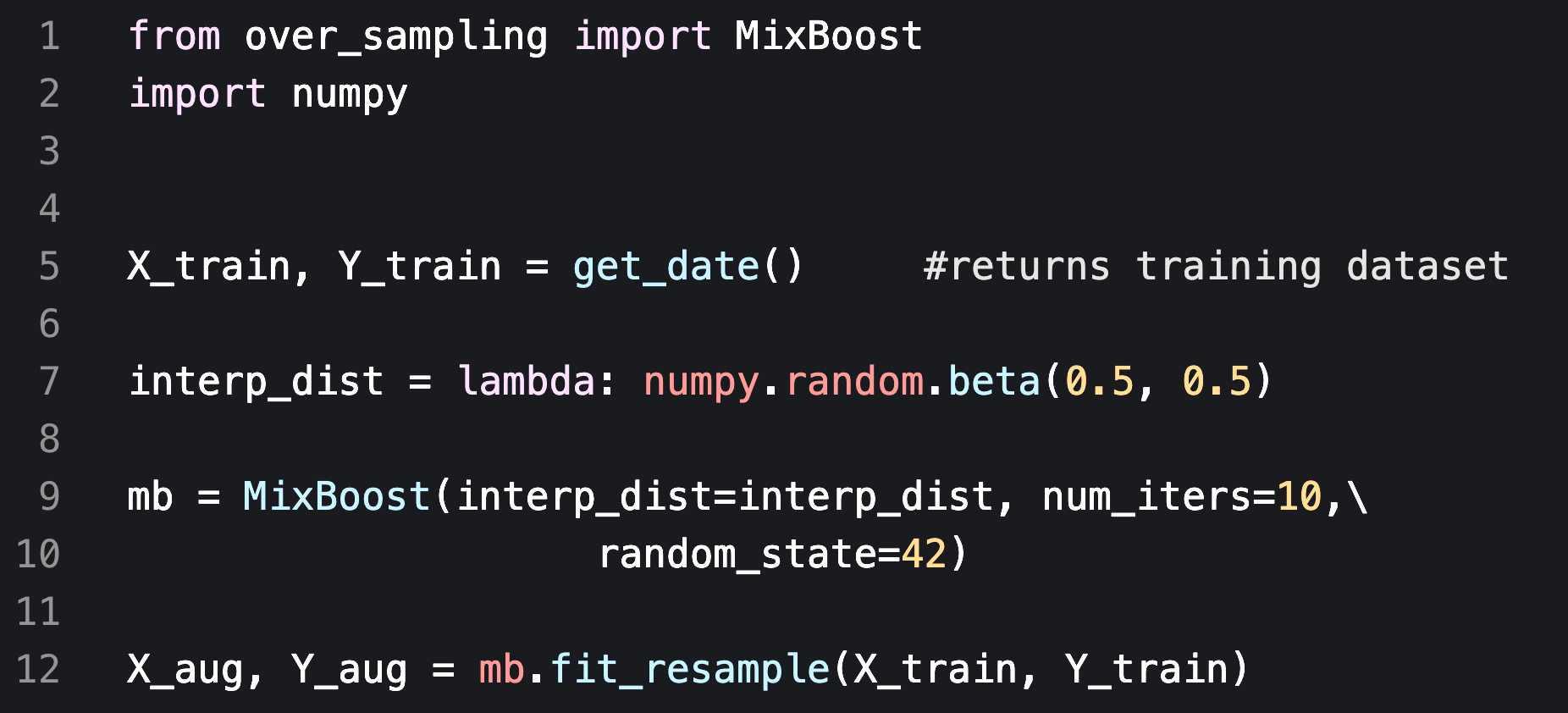}
    \caption{Code snippet that illustrates how to use $MixBoost$ in Python.}
    \label{fig:psuedo-code}
\end{figure}
\subsection{Datasets}
To evaluate \OURNAME against the state-of-the-art oversampling methods, we compare performance on binary classification on \textbf{20 benchmark datasets} (Table~\ref{tab:dataset-statistics}, available here\footnote{Datasets can be downloaded from~\cite{uci_archive}}).
To ensure the evaluation consistency we use the same dataset and imbalance ratios as in~\citet{swim}. 
For robust evaluation, we split the dataset into equal training and testing halves.
We randomly down-sample the minority class instances in the training dataset to simulate different levels of extreme imbalance.
Specifically, we test at three levels of imbalance, with minority training dataset sizes of 4, 7, and 10 to simulate the extreme imbalance scenarios~\cite{swim}.
We further skew the data distribution (2 \& 3 minority class instances) to show that \OURNAME outperforms existing methods when using even fewer number of minority class training instances.

\subsection{Classification}
For all experiments, we use the Naive Bayes, Nearest Neighbour, Decision Tree, Support Vector Machine, and Multi-Layer Perceptron (MLP) binary classifiers. 
For each dataset, we compare the best performance of each data augmentation method across the various binary classifiers against the performance of \OURNAME. 
% Since \OURNAME generates hybrid instances, we use only the MLP classifier to generate evaluation scores for \OURNAME. 
We use the \textit{sklearn}~\cite{pedregosa2011scikit} package in Python for our experiments; all classifiers use the default parameters.
We use \textit{tensorflow} to implement the MLP. 
For the MLP, we use \textit{Adam} optimizer, a learning rate of $1e-4$, a batch size of 500, and train it for 300 epochs. We use the same hyper-parameters for all datasets.
We normalize the data before training and testing.

\subsection{Data Augmentation}
We use each data augmentation method to generate $n$ synthetic instances, where $n$ is the number of instances in the training dataset.
For $MixBoost$, we find that generating the $n$ instances over 5 iterations ($0.2n$ in each iteration) leads to the best results.
We sample the interpolation ratio $\lambda$ from a $Beta(0.5, 0.5)$ distribution. 
We explore the sensitivity of $MixBoost$ to these hyper-parameters in Section~\ref{sec:abalation-studies}.

\subsection{Evaluation}
\label{sec:experimental-settings:evaluation}
We compare data augmentation methods on two metrics. 
First, we use \gmean \cite{kubat1997addressing}, which is the geometric mean of the True Positive Rate ($TPR$) (for majority class) and the True Negative Rate ($TNR$) (for minority class).
\begin{equation}
    \gmean = \sqrt{TPR~\times~TNR}
\end{equation}
We use \gmean because it is both immune to imbalance~\cite{kubat1997addressing} and provides information about extreme imbalance~\cite{swim}.
Second, for completeness, we also compare methods using the ROC-AUC scores.
We repeat all experiments 30 times and report the mean and standard deviation (rounded to the nearest decimal place) of the metrics for each dataset. Figure~\ref{fig:psuedo-code} includes a code snippet illustrating \OURNAME usage.

We compare \OURNAME to Random Over-sampling and Under-sampling (ROS, RUS), SMOTE~\cite{chawla2002smote}, Borderline-SMOTE (B1, B2)~\cite{borderlinesmote}, SMOTE with Tomek Links~\cite{kubat1997addressing}, ADASYN~\cite{adasyn} and SWIM~\cite{swim}. 
To make reporting easier, we merge all re-sampling approaches (ROS, RUS, SMOTE, B1, B2, SMOTE with Tomek Links, ADASYN) as Alternative Re-sampling Techniques (ALT) and report the best performing combination of data augmentation method and classifier for each dataset. 
We report the results from SWIM \cite{swim} separately since it is the most recent work, and it outperforms methods in ALT on most datasets. 
Finally, $Baseline$ represents the case where no data augmentation is used before training the binary classifier. 

\begin{table}
    \centering
    \small
    \begin{tabular}{|p{0.109\textwidth}|p{0.061\textwidth}|p{0.075\textwidth}|c|c|c|}
        \hline
        \multirow{3}{*}{\centering\textbf{Dataset}} & \multirow{3}{*}{\centering\textbf{Features}} & \multirow{3}{*}{\parbox{0.075\textwidth}{\centering\textbf{No. of majority instances}}} & \multirow{3}{*}{\centering\textbf{R4}} & \multirow{3}{*}{\centering\textbf{R7}} & \multirow{3}{*}{\centering\textbf{R10}} \\
        & & & & & \\
        & & & & & \\ \hline
        Abalone 9-18 & 8 & 689 & 1:173 & 1:99 & 1:69 \\
        Diabetes & 8 & 500 & 1:125 & 1:72 & 1:50 \\
        Wisconsin & 9 & 444 & 1:111 & 1:64 & 1:456 \\
        Wine Q. Red 4 & 11 & 1546 & 1:387 & 1:221 & 1:155 \\
        Wine Q. White & 11 & 880 & 1:220 & 1:126 & 1:88 \\
        Vowel 10 & 13 & 898 & 1:225 & 1:129 & 1:90 \\
        Pima Indians & 8 & 500 & 1:125 & 1:72 & 1:50 \\
        Vehicle 0 & 18 & 641 & 1:160  & 1:91 & 1:64 \\
        Vehicle 1 & 18 & 624 & 1:156  & 1:89 & 1:62\\
        Vehicle 2 & 18 & 622 & 1:155 & 1:88 & 1:62\\
        Vehicle 3 & 18 & 627 & 1:156 & 1:89 & 1:62\\
        Ring Norm & 20 & 3736 & 1:934 & 1:534 & 1:374 \\
        Waveform & 21 & 600 & 1:150 & 1:86 & 1:60 \\
        PC4 & 37 & 1280 & 1:320 & 1:183 & 1:128 \\
        Piechart & 37 & 644 & 1:161 & 1:92 & 1:65 \\
        Pizza Cutter & 37 & 609 & 1:153 & 1:87 & 1:61 \\
        Ada Agnostic & 48 & 3430 & 1:858 & 1:490 & 1:343 \\
        Forest Cover & 54 & 2970 & 1:743 & 1:425 & 1:297 \\
        Spam Base & 57 & 2788 & 1:697 & 1:399 & 1:279 \\
        Mfeat Karhu. & 64 & 1800 & 1:450 & 1:258 & 1:180 \\
        \hline
    \end{tabular}
    \caption{Description of the datasets used in our experiments. 
    To ensure evaluation consistency, we use the same datasets and configuration as proposed by \citet{swim}. 
    R4, R7, and R10 denote the ratio of class imbalance (minority:majority) after down-sampling the training datasets to have 4, 7, and 10 minority class instances respectively to simulate the extreme imbalance~\cite{swim} scenarios (as discussed in Section~\ref{sec:introduction})
    }
    \label{tab:dataset-statistics}
\end{table}
\begin{table}[]
    \centering
    \normalsize
    \begin{tabular}{|l|r|r|r|r|}
        \hline
        \textbf{Dataset} & \textbf{Baseline} & \textbf{ALT} & \textbf{SWIM} & \textbf{\OURNAME} \\ \hline
        Abalone 9-18 & 0.481 & 0.612 & 0.723 & \textbf{0.743} \\
        Diabetes & 0.259 & 0.509 & 0.509 & \textbf{0.701} \\
        Wisconsin & 0.874 & 0.956 & 0.958 & \textbf{0.969} \\
        Wine Q. Red 4 & 0.224 & 0.502 & 0.535 & \textbf{0.815} \\
        Wine Q.White 3v7 & 0.451 & 0.572 & 0.730 & \textbf{0.750} \\
        Vowel 10 & 0.724 & 0.738 & 0.812 & \textbf{0.845} \\
        Pima Indians & 0.276 & 0.479 & 0.509 & \textbf{0.700} \\
        Vehicle 0 & 0.534 & 0.758 & 0.814 & \textbf{0.900} \\
        Vehicle 1 & 0.541 & 0.739 & \textbf{0.791} & 0.735 \\
        Vehicle 2 & 0.450 & 0.549 & 0.560 & \textbf{0.880} \\
        Vehicle 3 & 0.402 & 0.505 & 0.569 & \textbf{0.651} \\
        Ring Norm & 0.274 & \textbf{0.933} & 0.899 & 0.580 \\
        Waveform & 0.301 & 0.701 & 0.688 & \textbf{0.844} \\
        PC4 & 0.572 & 0.559 & 0.611 & \textbf{0.737} \\
        PieChart & 0.455 & 0.516 & 0.576 & \textbf{0.741} \\
        Pizza Cutter & 0.468 & 0.506 & 0.552 & \textbf{0.725} \\
        Ada Agnostic & 0.451 & 0.445 & 0.539 & \textbf{0.690} \\
        Forest Cover & 0.561 & 0.554 & 0.550 & \textbf{0.917} \\
        Spam Base & 0.440 & 0.550 & 0.685 & \textbf{0.872} \\
        Mfeat Karhunen & 0.274 & \textbf{0.933} & 0.899 & 0.927 \\ \hline
    \end{tabular}
    \caption{\label{tab:results}%
    Comparative \gmean results (mean) for \OURNAME with existing over-sampling methods.
    These results represent the R4 setting where the training dataset has 4 minority class instances.
    Baseline refers to the case where classifier is trained without data augmentation.
    ALT is the score of the best performing data augmentation strategy (other than SWIM~\cite{swim} and \OURNAME) as described in Section~\ref{sec:experimental-settings:evaluation}.
    The best score for each dataset is highlighted in \textbf{bold}.}
    \label{tab:results-gmean}
\end{table}

\section{Results}
\label{results}
In this section, we present quantitative analysis of the proposed \OURNAME oversampling technique. 
First, we compare \OURNAME with existing state-of-the-art techniques and report results using the \gmean metric (Table~\ref{tab:results-gmean}). Next, for completeness of study, we report comparative performance on the ROC-AUC metric (Table~\ref{tab:results-roc}), contrast the different \OURNAME instance selection strategies (Table~\ref{tab:results-gmean-MixBoost}) and perform significance tests for different levels of imbalance (Section~\ref{sec:significance-testing}).

\subsection{Quantitative Results}
As in previous works, we compare \OURNAME with existing state-of-the-art techniques in Table~\ref{tab:results-gmean} using the \gmean metric, which is considered a better metric for evaluating performance when dealing with  extreme imbalance in datasets~\cite{swim,kubat1997addressing}. The results shown here are obtained by down-sampling to the most extreme imbalance setting R4 where there are only 4 minority class instances (see Table ~\ref{tab:dataset-statistics}). As evident, 
\OURNAME outperforms the existing methods on \textbf{17 out of the 20 datasets}. SWIM achieves best performance on only one while ALT achieves superior performance on 2 datasets. For each dataset, ALT reports the best performance amongst all algorithms other than SWIM and \OURNAME which are stated in \ref{sec:experimental-settings:evaluation}. In \OURNAME, the best results of the \textit{R-Selection} and \textit{EW-Selection} strategies are reported on each dataset.
\begin{table}[!h]
    \centering
    \normalsize
    \begin{tabular}{|l|r|r|}
        \hline
        \multirow{2}{*}{\centering\textbf{Dataset}} & \multicolumn{2}{c|}{\textbf{\OURNAME}} \\ \cline{2-3}
         & \textbf{R-Selection} & \textbf{EW-Selection} \\ \hline
        Abalone 9-18 & 0.743 $\pm$ 0.03 & 0.735 $\pm$ 0.06 \\
        Diabetes & 0.701 $\pm$ 0.05 & 0.560 $\pm$ 0.04 \\
        Wisconsin & 0.960 $\pm$ 0.02 & 0.969 $\pm$ 0.08 \\
        Wine Q. Red 4 & 0.714 $\pm$ 0.04 & 0.815 $\pm$ 0.08 \\
        Wine Q. White 3v7 & 0.743 $\pm$ 0.06 & 0.750 $\pm$ 0.05 \\
        Vowel 10 & 0.845 $\pm$ 0.04 & 0.854 $\pm$ 0.07 \\
        Pima Indians & 0.700 $\pm$ 0.05 & 0.597 $\pm$ 0.04 \\
        Vehicle 0 & 0.900 $\pm$ 0.05 & 0.850 $\pm$ 0.02 \\
        Vehicle 1 & 0.700 $\pm$ 0.03 & 0.735 $\pm$ 0.03 \\
        Vehicle 2 & 0.880 $\pm$ 0.02 & 0.638 $\pm$ 0.03 \\
        Vehicle 3 & 0.651 $\pm$ 0.06 & 0.600 $\pm$ 0.03 \\
        Ring Norm & 0.550 $\pm$ 0.04 & 0.580 $\pm$ 0.03 \\
        Waveform & 0.812 $\pm$ 0.03 & 0.844 $\pm$ 0.05 \\
        PC4 & 0.720 $\pm$ 0.08 & 0.737 $\pm$ 0.04 \\
        PieChart & 0.611 $\pm$ 0.06 & 0.741 $\pm$ 0.07 \\
        Pizza Cutter & 0.725 $\pm$ 0.05 & 0.678 $\pm$ 0.07 \\
        Ada Agnostic & 0.690 $\pm$ 0.02 & 0.648 $\pm$ 0.03 \\
        Forest Cover & 0.910 $\pm$ 0.05 & 0.917 $\pm$ 0.02 \\
        Spam Base & 0.872 $\pm$ 0.03 & 0.834 $\pm$ 0.05 \\
        Mfeat Karhunen & 0.888 $\pm$ 0.07 & 0.927 $\pm$ 0.05 \\ \hline
    \end{tabular}
    \caption{\label{tab:results-MixBoost}%
    Comparative \gmean results (mean and standard deviation for 30 independent runs) for the different candidate selection strategies (refer to Section~\ref{results} for details) of our approach \OURNAME.
    }
    \label{tab:results-gmean-MixBoost}
\end{table}
\begin{table}[!b]
    \centering
    \normalsize
    \begin{tabular}{|l|l|l|}
        \hline
        \textbf{Dataset} & \textbf{SWIM} & \textbf{\OURNAME} \\ \hline 
        Abalone 9-18 & 0.790 $\pm$ 0.08 & \textbf{0.820} $\pm$ 0.06 \\
        Diabetes & 0.778 $\pm$ 0.03 & \textbf{0.800} $\pm$ 0.02 \\
        Wisconsin & 0.899 $\pm$ 0.08 & \textbf{0.972} $\pm$ 0.01 \\
        Wine Q. Red 4 & 0.950 $\pm$ 0.03 & \textbf{0.971} $\pm$ 0.03 \\
        Wine Q. White 3 vs 7 & 0.640 $\pm$ 0.09 & \textbf{0.783} $\pm$ 0.05 \\
        Vowel 10 & 0.895 $\pm$ 0.05 & \textbf{0.958} $\pm$ 0.03 \\
        Pima Indians & 0.778 $\pm$ 0.03 & \textbf{0.800} $\pm$ 0.02 \\
        Vehicle 0 & 0.628 $\pm$ 0.04 & \textbf{0.896} $\pm$ 0.07 \\
        Vehicle 1 & 0.592 $\pm$ 0.06 & \textbf{0.750} $\pm$ 0.02 \\
        Vehicle 2 & 0.824 $\pm$ 0.03 & \textbf{0.921} $\pm$ 0.03 \\
        Vehicle 3 & 0.585 $\pm$ 0.05 & \textbf{0.664} $\pm$ 0.03 \\
        Ring Norm & 0.704 $\pm$ 0.24 & \textbf{0.892} $\pm$ 0.08 \\
        Waveform & 0.828 $\pm$ 0.02 & \textbf{0.869} $\pm$ 0.02 \\
        PC4 & 0.686 $\pm$ 0.09 & \textbf{0.794} $\pm$ 0.03 \\
        PieChart & 0.661 $\pm$ 0.06 & \textbf{0.743} $\pm$ 0.08 \\
        Pizza Cutter & 0.662 $\pm$ 0.08 & \textbf{0.735} $\pm$ 0.04 \\
        Ada Agnostic & 0.671 $\pm$ 0.05 & \textbf{0.798} $\pm$ 0.02 \\
        Forest Cover & 0.881 $\pm$ 0.07 & \textbf{0.970} $\pm$ 0.02 \\
        Spam Base & 0.769 $\pm$ 0.12 & \textbf{0.835} $\pm$ 0.07 \\
        Mfeat Karhunen & 0.980 $\pm$ 0.01 & \textbf{0.992} $\pm$ 0.00 \\ \hline
    \end{tabular}
    \caption{
    ROC-AUC scores (mean and standard deviation) for EW-Selection strategy of \OURNAME and SWIM. The best score for each dataset is highlighted in \textbf{bold}.
    }%
    \label{tab:results-roc}
\end{table}
Table~\ref{tab:results-gmean-MixBoost} compares the performance of the different selection strategies used in \OURNAME. The relative performance of \textit{EW-Selection} and \textit{R-Selection} strategies for $\OURNAME$ indicates that leveraging an entropy prior during the candidate selection leads to the creation of more useful synthetic hybrid candidates than selecting instances at random, in several cases. Interestingly, both \textit{R-Selection} and \textit{EW-Selection} strategies independently outperform existing methods on \textbf{17 out of 20 datasets}.

Table~\ref{tab:results-roc} reports comparative results with the ROC-AUC metric. For ease of analysis, we restrict the comparison to SWIM since it is the strongest baseline which consistently outperforms other techniques and is also the most recent state-of-the-art. The results highlight that \OURNAME outperforms SWIM on the ROC-AUC metric as well by achieving superior performance on \textbf{20 out of 20 datasets}. 

To study the sensitivity of \OURNAME, we analyse and compare standard deviation for different combinations of \OURNAME along with SWIM in Tables~\ref{tab:results-gmean-MixBoost} and ~\ref{tab:results-roc}. The results show that \OURNAME results in a mean standard deviation (over all 20 datasets) of 0.047 which is comparable with 0.044 by SWIM. In the next section, we also probe the statistical significance of our observed results, with respect to the strongest baseline (SWIM), across varying levels of class imbalance.
\begin{figure*}[t]
    \centering
    % \subfigure[text]{
    % \includegraphics[width=0.33\textwidth]{figures/download.png}
    % }
    % \subfigure[text]{
    % \includegraphics[width=0.33\textwidth]{figures/7samp.png}
    % }
    % \subfigure[text]{
    % \includegraphics[width=0.33\textwidth]{figures/10samp.png}
    % }
    \begin{subfigure}[t]{0.3\linewidth}
        \includegraphics[width=\linewidth,trim={0 0 0 0.7cm},clip]{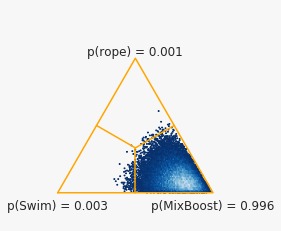}
        \caption{Figure A}
    \end{subfigure}\hfil%
    \begin{subfigure}[t]{0.3\linewidth}
        \includegraphics[width=\linewidth,trim={0 0 0 0.7cm},clip]{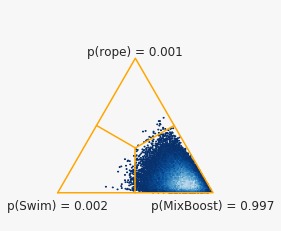}
        \caption{Figure B}
    \end{subfigure}\hfil%
    \begin{subfigure}[t]{0.3\linewidth}
        \includegraphics[width=\linewidth,trim={0 0 0 0.7cm},clip]{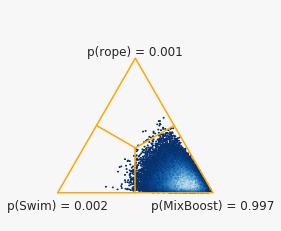}
        \caption{Figure C}
    \end{subfigure}%
    \caption{Posteriors for SWIM versus \OURNAME (left and right vertex) on the datasets with 4, 7 and 10 minority class instances ((a), (b), and (c)) for the Bayesian sign-rank test.
    A higher concentration of points on one of the sides of the triangle shows that a given method has a higher probability of being statistically significantly better~\cite{swim}.
    The top vertex $rope$ indicates the case where neither method is statistically significantly better than the other.}
    \label{fig:results:bayesian-testing}
\end{figure*}

\subsection{Statistical Significance}
\label{sec:significance-testing}
We use the Bayesian signed test~\cite{bayes_sign_test} to evaluate the results presented in the previous section. 
The Bayesian signed test is used to compare two classification methods over several datasets. 
We use the test to compare \OURNAME to SWIM over all 20 datasets. 
We compare the methods for training datasets down-sampled to 4, 7, and 10 minority class instances.
Using the Bayesian method enables us to ask questions about posterior probabilities, that we cannot answer using null hypothesis tests~\cite{swim}. 

Concretely, we wish to ascertain, based on the experiments, what is the probability that \OURNAME is better than SWIM for data augmentation. 
We repeat the setup described in \cite{swim}, where, based on the assumption of the Dirichlet process, the posterior probability for the Bayesian signed test is calculated as a mixture of Dirac deltas centered on the observation.

Figure~\ref{fig:results:bayesian-testing} shows the posterior plots of the Bayesian signed test for the three down-sampled datasets. 
The posteriors are calculated with the prior parameter of the Dirichlet as $s = 0.5$ and $z_{0} = 0$ as suggested by \cite{swim}. 
The posterior plots report the samples from the posterior (blue cloud of points), the simplex (the large triangle), and three regions. The three regions of the triangle denote the following,
(1) The region in the bottom left of the triangle indicates the case where it is more probable that SWIM is better than \OURNAME.
(2) The region in the bottom right of the triangle indicates the opposite, i.e., when it is more probable that \OURNAME is better than SWIM.
(3) The region at the top of the triangle indicates that it is likely that neither method is better.
The closer the points are to the sides of the triangle, the larger is the statistical difference between methods.
Figure~\ref{fig:results:bayesian-testing} shows that the probability that \OURNAME is better than SWIM for data augmentation is high for all three down-sampled training datasets. 
For all three plots, the point cloud is concentrated in the bottom right triangle, indicating that \OURNAME is statistically significantly better than SWIM.
\section{Ablation Studies}
\label{sec:abalation-studies}
In this section, we evaluate the specific impact of different components of \OURNAME and measure the sensitivity of our framework to hyper-parameters.

To balance the consistency of results and ease of analysis and presentation, we restrict our discussion to five datasets, namely Pima Indians, Waveform, PC4, Piechart and Forest Cover. These datasets are selected to ensure maximum \textit{diversity in feature dimensionality} (ranging from 8 features for \textit{Pima Indians} to 54 for \textit{Forest Cover}) and the \textit{extent of class imbalance} (ranging from 1:125 for \textit{Pima Indians} to 1:743 for \textit{Forest Cover}).

For each study (unless otherwise specified), we augment the training dataset with $n$ synthetic hybrid instances, where $n$ is the number of instances in the training dataset. As in main experiments, these synthetic hybrid instances are generated equally over \textit{five} \OURNAME iterations ($0.2n$ instances in each iteration). For the scope of this analysis, we use the \textit{EW-Selection} technique in 
\OURNAME and sample the interpolation ratio $\lambda$ from a $Beta(0.5, 0.5)$ distribution.

\subsection{Impact of sampling instances over multiple iterations}
\label{sec:discussion-iterative-sampling}
In this experiment, we evaluate the importance of generating synthetic hybrid instances over multiple iterations.
We define a variant of $\OURNAME$, called \textit{MixBoost-1-Iteration} (denoted by \textit{MixBoost-1-Iter}), where all the $n$ synthetic instances are generated in a \textit{single} iteration.
This single-step generation is in contrast to $\OURNAME$ used previously, where the $n$ instances are generated over \textit{five} iterations, $0.2n$ at each iteration.

We augment the training dataset with the generated instances and train the binary classifier on the augmented dataset. 
Table~\ref{table:iterative-vs-1shot} shows the results. 
We observe that the sampling instances over multiple (\textit{five}) iterations in $\OURNAME$ outperforms the single-step variant on each datasets. This result corroborates the importance of the iterative characteristic ($Boost$ step) of $\OURNAME$. 

\begin{table}[h]
    \centering
    \normalsize
    \begin{tabular}{|l|r|r|}
        \hline
        \textbf{Dataset} & \textbf{\textit{MixBoost-1-Iter}} & \textbf{\OURNAME}           \\ \hline
        Pima Indians     & 0.491 $\pm$ 0.02 & \textbf{0.597 $\pm$ 0.04} \\
        Waveform         & 0.843 $\pm$ 0.04 & \textbf{0.844 $\pm$ 0.05} \\
        PC4              & 0.613 $\pm$ 0.08 & \textbf{0.737 $\pm$ 0.04} \\
        Piechart         & 0.721 $\pm$ 0.02 & \textbf{0.741 $\pm$ 0.07} \\
        Forest Cover     & 0.910 $\pm$ 0.00 & \textbf{0.917 $\pm$ 0.02} \\ \hline
    \end{tabular}
    \caption{\gmean scores for the single step (\OURNAME-1-Iter) and the proposed $\OURNAME$. For all selected datasets, the iterative variant of $\OURNAME$ outperforms the single step one.
    }
    \label{table:iterative-vs-1shot}
\end{table}
% \begin{table}[h]
%     \begin{tabular}{|l|r|r|}
%     \hline
%     \textbf{Dataset}  & \textbf{MixBoost\-1\-Iteration} & \textbf{\OURNAME} \\ \hline
%     Pima Indians    & 0.510 & \textbf{0.597} \\
%     Waveform        & 0.841 & \textbf{0.844} \\
%     PC4             & 0.630 & \textbf{0.737} \\
%     Piechart        & 0.700 & \textbf{0.741} \\
%     Forest Cover    & 0.914 & \textbf{0.917} \\ 
%     \hline
%     \end{tabular}
%     \caption{\gmean scores (mean and standard deviation over 30 runs) for the single step and iterative variants of $\OURNAME$. 
%     For all selected datasets, the iterative variant of $\OURNAME$ outperforms the single step one. 
%     }
%     \label{tab:iterative-vs-1shot}
% \end{table}

\subsection{Impact of choice of  distributions for sampling $\lambda$}
\label{sec:discussion-lambda-sampling}
For sampling the interpolation ratio, $\lambda$, we experiment with Linear and Non-Linear Probability Density Functions (PDFs).
Table~\ref{tab:comparison-with-different-lambda-distributions} shows the results. 
On all datasets, using a non-linear PDF ($Beta(\alpha,\alpha)$ with $\alpha=0.5$) in the $Mix$ step leads to better classifier performance.
\begin{table}[h]
    \centering
    \normalsize
    \begin{tabular}{|l|r|r|}
        \hline
        \textbf{Dataset} & \multicolumn{1`}{c|}{\textbf{Uniform}}  & \multicolumn{1}{c|}{\textbf{Beta}} \\ \hline 
        Pima Indians & 0.389 $\pm$ 0.08  & \textbf{0.597 $\pm$ 0.04} \\ 
        Waveform     & 0.661 $\pm$ 0.01 & \textbf{0.844 $\pm$ 0.05}   \\
        PC4          & 0.242 $\pm$ 0.01 & \textbf{0.737 $\pm$ 0.04}   \\
        Piechart     & 0.312 $\pm$ 0.07  & \textbf{0.741 $\pm$ 0.07}    \\
        Forest Cover & 0.629 $\pm$ 0.01  & \textbf{0.917 $\pm$ 0.02} \\
        \hline
    \end{tabular}
    \caption{\gmean scores when we sample $\lambda$ from different distributions.
    For all datasets, sampling $\lambda$ from the \textit{Beta(0.5,0.5)} distribution leads to the best performance.}
    \label{tab:comparison-with-different-lambda-distributions}
\end{table}

\subsection{Impact of generating hybrid-instances}
\label{sec:discussion-weighted-crossentropy}
Existing approaches augment training datasets by oversampling the minority class instances. \OURNAME generates synthetic hybrid instances that have elements of both the minority and majority class. Concretely, in the $Mix$ step, we generate instances where the target class vector is in-between the majority class vector $[1, 0]$ and the minority class vector $[0, 1]$. Alternatively, in the $Mix$ step, we could have generated synthetic instances where the target class vector corresponded to either the majority or the minority class. To evaluate these alternatives, we create a variant of \OURNAME in which we assign either the majority class vector (when $\lambda > 0.5$) or the minority class vector (when $\lambda <= 0.5$) to the generated instance.
We label this variant \textit{MixBoost-OneHot}.
\begin{table}[h]
    \centering
    \normalsize
    \begin{tabular}{|l|r|r|}
        \hline
        \textbf{Dataset} & \textbf{\textit{MixBoost-OneHot}} & \multicolumn{1}{c|}{\textbf{\OURNAME}}           \\ \hline
        Pima Indians     & 0.458 $\pm$ 0.03        & \textbf{0.597 $\pm$ 0.04} \\
        Waveform         & 0.831 $\pm$ 0.01        & \textbf{0.844 $\pm$ 0.05}   \\
        PC4              & 0.698 $\pm$ 0.07        & \textbf{0.737 $\pm$ 0.04}   \\
        Piechart         & 0.566 $\pm$ 0.06        & \textbf{0.741 $\pm$ 0.07}    \\
        Forest Cover     & 0.898 $\pm$ 0.02        & \textbf{0.917 $\pm$ 0.02}   \\ \hline
    \end{tabular}
    \caption{\gmean scores comparing \OURNAME to a variant where we generate non-hybrid instances (\textit{MixBoost-OneHot}) in the $Mix$ step. For all considered datasets, generating hybrid instances improves the performance of the binary classifier.}
    \label{tab:onehot-vs-notonehot}
\end{table}
Table~\ref{tab:onehot-vs-notonehot} shows the results comparing \OURNAME to \textit{MixBoost-OneHot}.
We observe that generating hybrid instances with in-between target class labels (from \OURNAME) outperforms generating instances with either a majority or a minority class label (from \textit{MixBoost-OneHot}).  
We posit that training with hybrid instances that are not explicitly attributed to either class enables the learning of more robust decision boundaries by helping regulate the confidence of the classifier away from the training distribution. 

\subsection{Impact of decreasing number of minority class instances}
\label{sec:discussion-num-samples}
Since $\OURNAME$ uses an intelligent selection technique for generating synthetic hybrid instances, we want to study whether it outperforms existing approaches with a fewer number of minority class training instances.
We generate variations of the training dataset with 2 and 3 minority class instances. 
We then run $\OURNAME$ on these variations and train a classifier on the augmented dataset.  
Table~\ref{table:ComparisonWithNoOfSamples} shows the results. 
$\OURNAME$ achieves better results than SWIM while using a fewer number of minority class training examples. For instance, for the Waveform and Pima Indians dataset, $\OURNAME$ outperforms SWIM using half the number of minority class instances (2 vs. 4). 
\begin{table*}[!th]
    \centering
    % \footnotesize
    \normalsize
    \begin{tabular}{|l|r|r|r|r|}
    \hline
        \multirow{2}{*}{\textbf{Dataset}} & \multicolumn{1}{c|}{\textbf{SWIM}} & \multicolumn{3}{c|}{\textbf{\OURNAME}} \\ \cline{2-5} 
         & $min_{ct}=4$ & $min_{ct}=2$ & $min_{ct}=3$ & $min_{ct}=4$ \\ \hline
        Pima Indians & 0.499 $\pm$ 0.13 & 0.534 $\pm$ 0.08 & 0.616  $\pm$ 0.02 & \textbf{0.597 $\pm$ 0.04} \\
        Waveform & 0.652 $\pm$ 0.03 & 0.693 $\pm$ 0.06 & 0.727 $\pm$ 0.03 & \textbf{0.844 $\pm$ 0.05} \\
        PC4 & 0.661 $\pm$ 0.06 & 0.593 $\pm$ 0.06 & 0.688 $\pm$ 0.02 & \textbf{0.737 $\pm$ 0.04} \\
        PieChart & 0.612 $\pm$ 0.05 & 0.533 $\pm$ 0.13 & 0.579 $\pm$ 0.09 & \textbf{0.741 $\pm$ 0.07} \\
        Forest Cover & 0.522 $\pm$ 0.03 & 0.452 $\pm$ 0.05 & 0.643 $\pm$ 0.03 & \textbf{0.917} $\pm$ \textbf{0.02} \\ \hline
    \end{tabular}
    \caption{\gmean scores for data augmentation using \OURNAME and SWIM as we reduce the number of minority class instances ($min_{ct}$) by down-sampling. \OURNAME achieves best performance on all the datasets. Additionally, \OURNAME outperforms SWIM on 4/5 datasets when using fewer (3 vs 4) the minority class instances than SWIM.}
    \label{table:ComparisonWithNoOfSamples}
\end{table*}

\subsection{Impact of number of generated synthetic hybrid-instances}
\label{sec:discussion-IEW-over-sampling-steps}
\begin{figure}[!h]
    \includegraphics[width=0.4\textwidth]{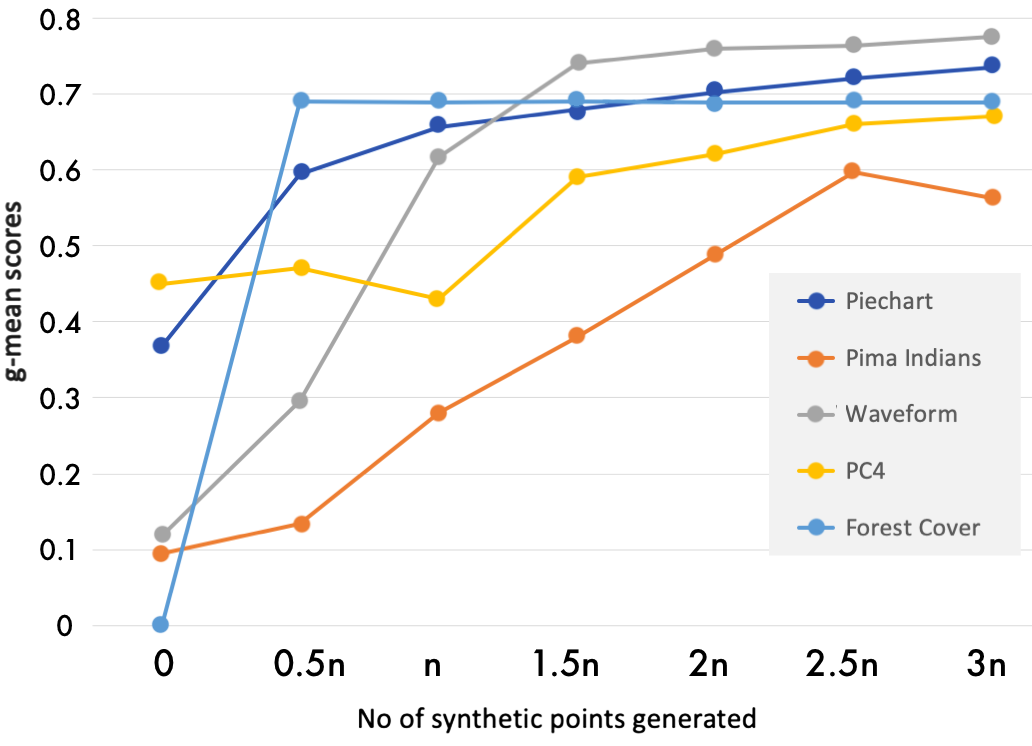}
    \centering
    \caption{Variation in \gmean scores (averaged over 30 runs) for \OURNAME as we increase the number of generated synthetic hybrid instances. 
    $n$ represents the total number of training instances in the original dataset.
    The different color lines indicate different datasets.
    The gain of generating an additional $0.5n$ synthetic instances is initially high and then falls gradually as the number of generated instances increases. 
    }
    \label{fig:gscore-vs-iterations}
\end{figure}

\OURNAME generates synthetic instances iteratively. 
For each iteration, the classifier is retrained using all instances generated up to this iteration. 
We expect that the marginal gain of generating instances should decrease as more instances are generated. 
Figure~\ref{fig:gscore-vs-iterations} validates our expectation. 
We see that the performance of the classifier plateaus as \OURNAME adds more synthetic instances to the training dataset.
This plateau in performance is important for the practical deployment of \OURNAME.

\subsection{Impact of number of minority class instances}
\label{sec:discussion-inc-min-samples}
\begin{figure}[h]
    \includegraphics[width=0.47\textwidth]{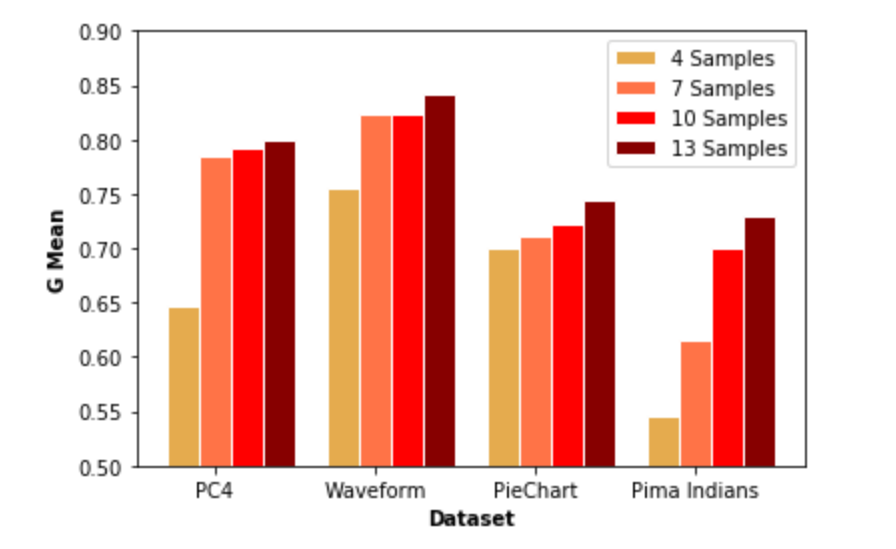}
    \centering
    \caption{Variation in \gmean scores (averaged over 30 runs) for different number of minority class instances using \OURNAME. For each dataset, the classifier performance increases with an increase in the number of minority class instances.
    }
    \label{fig:gscore-vs-minority-sampled}
\end{figure}

\OURNAME mixes an instance of the minority class with an instance of the majority class to create a synthetic hybrid instance.
We expect the quality of the generated synthetic instances to improve as we increase the number of minority class instances in the training dataset. The quality of generated instances is measured by classifier performance. Figure \ref{fig:gscore-vs-minority-sampled} shows that the classifier performance improves as we increase the number of minority class instances in the training data. 

\section{Related Work}
\label{sec:related-work}
There are broadly two categories of approaches for dealing with classification problems on imbalanced datasets. 
First, sampling-based approaches and second, cost-based approaches~\cite{elkan2001foundations,wang2016fast}. 
In this paper, we focus on sampling-based approaches. 
The most straightforward re-sampling strategies are Random under Sampling (RUS), and Random over Sampling (ROS). 
They balance the class distribution by either randomly deleting instances of the majority class (RUS) or randomly duplicating instances of the minority class (ROS). 
However, deleting instances from the training dataset can lead to a loss of information. 
Further, duplicating instances of the minority class does not add any new information. 

SMOTE (Synthetic Minority Oversampling Technique)~\cite{chawla2002smote} alleviates these shortcomings. 
SMOTE balances the class distribution by generating synthetic data instances.
SMOTE generates a synthetic instance by interpolating the k-nearest neighbors of a minority class instance in the training data. 
SMOTE has proved useful in a variety of contexts~\cite{chawla2002smote,chawla2003smoteboost,borderlinesmote}.
However, since SMOTE generates synthetic instances using only minority class instances, the generated instances lie within the convex hull of the minority class distribution.
Further, because it does not use majority class instances, SMOTE can increase the overlap between classes. 
Therefore, if the classes are extremely imbalanced, then SMOTE can degrade classifier performance. 
Extensions of SMOTE~\cite{fernandez2018smote} add a post-processing step that tries to remove generated instances that might degrade the performance of the classifier. 
These methods incorporate the majority class distribution into the post-processing step of the data augmentation process. 
For instance, Adaptive Synthetic Oversampling (ADASYN)~\cite{adasyn}, borderline SMOTE~\cite{borderlinesmote}, and Majority Weighted Minority Oversampling~\cite{he2008adaptive} use majority class instances present in the neighborhood of generated instances for post-processing. 
However, these methods use only the minority class distribution when generating synthetic instances. 
Therefore, if the number of minority class instances is very small, then the quality of generated instances and subsequently, the performance of the classifier is degraded. \citet{abdi2015combat} create a variant of SMOTE by using the Mahalanobis distance (instead of Euclidean distance) for generating synthetic instances. 
However, as with SMOTE, they generate instances using only minority class data.
SWIM~\cite{swim} uses information from the majority class to generate synthetic instances and the authors show the technique to outperforms previous methods across several benchmark datasets at extreme levels of class imbalance. 

\citet{gan1} approach the problem of generating synthetic instances from a different direction. 
They use a Generative Adversarial Network (GAN) trained on minority class data to generate synthetic training instances. 
However, their method is unable to consistently outperform existing state-of-the-art methods(\citet{gan1}, see results). 
Further, training a GAN for each dataset requires significant compute and hyper-parameter tuning. 
For these reasons, we have not compared \OURNAME to their method.

\section{Conclusion}
In the present work, we tackle the problem of binary classification on extremely imbalanced datasets. 
To this end, we propose \OURNAME, a technique for synthetic iterative oversampling. \OURNAME intelligently selects and then combines instances from the majority and minority classes to generate synthetic hybrid instance. We show that \OURNAME outperforms existing methods on several benchmark datasets. We also evaluate the impact of the different components of \OURNAME using ablation studies. As directions of future study, we wish to focus on evaluating \OURNAME for multi-class classification and adapt the idea of iterative sampling and generation through interleaved $Mix$ and $Boost$ steps for regression tasks.

% \bibliographystyle{IEEEtran}
% \bibliography{references}

\begin{thebibliography}{30}

%%% the Web address.

\bibitem[\protect\citeauthoryear{Abdi and Hashemi}{Abdi and Hashemi}{2015}]%
        {abdi2015combat}
\bibfield{author}{\bibinfo{person}{Lida Abdi} {and} \bibinfo{person}{Sattar
  Hashemi}.} \bibinfo{year}{2015}\natexlab{}.
\newblock \showarticletitle{To combat multi-class imbalanced problems by means
  of over-sampling techniques}.
\newblock \bibinfo{journal}{\emph{IEEE transactions on Knowledge and Data
  Engineering}} \bibinfo{volume}{28}, \bibinfo{number}{1}
  (\bibinfo{year}{2015}), \bibinfo{pages}{238--251}.
\newblock


\bibitem[\protect\citeauthoryear{Al-Shahib, Breitling, and Gilbert}{Al-Shahib
  et~al\mbox{.}}{2005}]%
        {al2005feature}
\bibfield{author}{\bibinfo{person}{Ali Al-Shahib}, \bibinfo{person}{Rainer
  Breitling}, {and} \bibinfo{person}{David Gilbert}.}
  \bibinfo{year}{2005}\natexlab{}.
\newblock \showarticletitle{Feature selection and the class imbalance problem
  in predicting protein function from sequence}.
\newblock \bibinfo{journal}{\emph{Applied Bioinformatics}} \bibinfo{volume}{4},
  \bibinfo{number}{3} (\bibinfo{year}{2005}), \bibinfo{pages}{195--203}.
\newblock


\bibitem[\protect\citeauthoryear{Benavoli, Corani, Mangili, Zaffalon, and
  Ruggeri}{Benavoli et~al\mbox{.}}{2014}]%
        {bayes_sign_test}
\bibfield{author}{\bibinfo{person}{Alessio Benavoli}, \bibinfo{person}{Giorgio
  Corani}, \bibinfo{person}{Francesca Mangili}, \bibinfo{person}{Marco
  Zaffalon}, {and} \bibinfo{person}{Fabrizio Ruggeri}.}
  \bibinfo{year}{2014}\natexlab{}.
\newblock \showarticletitle{A Bayesian Wilcoxon signed-rank test based on the
  Dirichlet process.}. In \bibinfo{booktitle}{\emph{ICML}}
  \emph{(\bibinfo{series}{JMLR Workshop and Conference Proceedings})},
  Vol.~\bibinfo{volume}{32}. \bibinfo{publisher}{JMLR.org},
  \bibinfo{pages}{1026--1034}.
\newblock
\urldef\tempurl%
\url{http://dblp.uni-trier.de/db/conf/icml/icml2014.html#BenavoliCMZR14}
\showURL{%
\tempurl}


\bibitem[\protect\citeauthoryear{Bennin, Keung, Phannachitta, Monden, and
  Mensah}{Bennin et~al\mbox{.}}{2017}]%
        {bennin2017mahakil}
\bibfield{author}{\bibinfo{person}{Kwabena~Ebo Bennin}, \bibinfo{person}{Jacky
  Keung}, \bibinfo{person}{Passakorn Phannachitta}, \bibinfo{person}{Akito
  Monden}, {and} \bibinfo{person}{Solomon Mensah}.}
  \bibinfo{year}{2017}\natexlab{}.
\newblock \showarticletitle{Mahakil: Diversity based oversampling approach to
  alleviate the class imbalance issue in software defect prediction}.
\newblock \bibinfo{journal}{\emph{IEEE Transactions on Software Engineering}}
  \bibinfo{volume}{44}, \bibinfo{number}{6} (\bibinfo{year}{2017}),
  \bibinfo{pages}{534--550}.
\newblock


\bibitem[\protect\citeauthoryear{Burez and Van~den Poel}{Burez and Van~den
  Poel}{2009}]%
        {burez2009handling}
\bibfield{author}{\bibinfo{person}{Jonathan Burez} {and} \bibinfo{person}{Dirk
  Van~den Poel}.} \bibinfo{year}{2009}\natexlab{}.
\newblock \showarticletitle{Handling class imbalance in customer churn
  prediction}.
\newblock \bibinfo{journal}{\emph{Expert Systems with Applications}}
  \bibinfo{volume}{36}, \bibinfo{number}{3} (\bibinfo{year}{2009}),
  \bibinfo{pages}{4626--4636}.
\newblock


\bibitem[\protect\citeauthoryear{Chawla, Bowyer, Hall, and Kegelmeyer}{Chawla
  et~al\mbox{.}}{2002}]%
        {chawla2002smote}
\bibfield{author}{\bibinfo{person}{Nitesh~V Chawla}, \bibinfo{person}{Kevin~W
  Bowyer}, \bibinfo{person}{Lawrence~O Hall}, {and} \bibinfo{person}{W~Philip
  Kegelmeyer}.} \bibinfo{year}{2002}\natexlab{}.
\newblock \showarticletitle{SMOTE: synthetic minority over-sampling technique}.
\newblock \bibinfo{journal}{\emph{Journal of artificial intelligence research}}
   \bibinfo{volume}{16} (\bibinfo{year}{2002}), \bibinfo{pages}{321--357}.
\newblock


\bibitem[\protect\citeauthoryear{Chawla, Lazarevic, Hall, and Bowyer}{Chawla
  et~al\mbox{.}}{2003}]%
        {chawla2003smoteboost}
\bibfield{author}{\bibinfo{person}{Nitesh~V Chawla},
  \bibinfo{person}{Aleksandar Lazarevic}, \bibinfo{person}{Lawrence~O Hall},
  {and} \bibinfo{person}{Kevin~W Bowyer}.} \bibinfo{year}{2003}\natexlab{}.
\newblock \showarticletitle{SMOTEBoost: Improving prediction of the minority
  class in boosting}. In \bibinfo{booktitle}{\emph{European conference on
  principles of data mining and knowledge discovery}}. Springer,
  \bibinfo{pages}{107--119}.
\newblock


\bibitem[\protect\citeauthoryear{Code}{Code}{2019}]%
        {github_code}
\bibfield{author}{\bibinfo{person}{Anonymous~MixBoost Code}.}
  \bibinfo{year}{2019}\natexlab{}.
\newblock \bibinfo{title}{MixBoost-Code}.
\newblock
  \bibinfo{howpublished}{\url{https://github.com/user1332/MIXBOOST-code}}.
\newblock


\bibitem[\protect\citeauthoryear{Dua and Graff}{Dua and Graff}{2017}]%
        {uci_archive}
\bibfield{author}{\bibinfo{person}{Dheeru Dua} {and} \bibinfo{person}{Casey
  Graff}.} \bibinfo{year}{2017}\natexlab{}.
\newblock \bibinfo{title}{{UCI} Machine Learning Repository}.
\newblock
\newblock
\urldef\tempurl%
\url{http://archive.ics.uci.edu/ml}
\showURL{%
\tempurl}


\bibitem[\protect\citeauthoryear{Elkan}{Elkan}{2001}]%
        {elkan2001foundations}
\bibfield{author}{\bibinfo{person}{Charles Elkan}.}
  \bibinfo{year}{2001}\natexlab{}.
\newblock \showarticletitle{The foundations of cost-sensitive learning}. In
  \bibinfo{booktitle}{\emph{International joint conference on artificial
  intelligence}}, Vol.~\bibinfo{volume}{17}. Lawrence Erlbaum Associates Ltd,
  \bibinfo{pages}{973--978}.
\newblock


\bibitem[\protect\citeauthoryear{Fern{\'a}ndez, Garcia, Herrera, and
  Chawla}{Fern{\'a}ndez et~al\mbox{.}}{2018}]%
        {fernandez2018smote}
\bibfield{author}{\bibinfo{person}{Alberto Fern{\'a}ndez},
  \bibinfo{person}{Salvador Garcia}, \bibinfo{person}{Francisco Herrera}, {and}
  \bibinfo{person}{Nitesh~V Chawla}.} \bibinfo{year}{2018}\natexlab{}.
\newblock \showarticletitle{SMOTE for learning from imbalanced data: progress
  and challenges, marking the 15-year anniversary}.
\newblock \bibinfo{journal}{\emph{Journal of artificial intelligence research}}
   \bibinfo{volume}{61} (\bibinfo{year}{2018}), \bibinfo{pages}{863--905}.
\newblock


\bibitem[\protect\citeauthoryear{Galar, Fernandez, Barrenechea, Bustince, and
  Herrera}{Galar et~al\mbox{.}}{2011}]%
        {galar2011review}
\bibfield{author}{\bibinfo{person}{Mikel Galar}, \bibinfo{person}{Alberto
  Fernandez}, \bibinfo{person}{Edurne Barrenechea}, \bibinfo{person}{Humberto
  Bustince}, {and} \bibinfo{person}{Francisco Herrera}.}
  \bibinfo{year}{2011}\natexlab{}.
\newblock \showarticletitle{A review on ensembles for the class imbalance
  problem: bagging-, boosting-, and hybrid-based approaches}.
\newblock \bibinfo{journal}{\emph{IEEE Transactions on Systems, Man, and
  Cybernetics, Part C (Applications and Reviews)}} \bibinfo{volume}{42},
  \bibinfo{number}{4} (\bibinfo{year}{2011}), \bibinfo{pages}{463--484}.
\newblock


\bibitem[\protect\citeauthoryear{Garc{\'\i}a and Herrera}{Garc{\'\i}a and
  Herrera}{2009}]%
        {garcia2009evolutionary}
\bibfield{author}{\bibinfo{person}{Salvador Garc{\'\i}a} {and}
  \bibinfo{person}{Francisco Herrera}.} \bibinfo{year}{2009}\natexlab{}.
\newblock \showarticletitle{Evolutionary undersampling for classification with
  imbalanced datasets: Proposals and taxonomy}.
\newblock \bibinfo{journal}{\emph{Evolutionary computation}}
  \bibinfo{volume}{17}, \bibinfo{number}{3} (\bibinfo{year}{2009}),
  \bibinfo{pages}{275--306}.
\newblock


\bibitem[\protect\citeauthoryear{Han, Wang, and Mao}{Han
  et~al\mbox{.}}{2005a}]%
        {borderlinesmote}
\bibfield{author}{\bibinfo{person}{Hui Han}, \bibinfo{person}{Wen-Yuan Wang},
  {and} \bibinfo{person}{Bing-Huan Mao}.} \bibinfo{year}{2005}\natexlab{a}.
\newblock \showarticletitle{Borderline-SMOTE: a new over-sampling method in
  imbalanced data sets learning}. In \bibinfo{booktitle}{\emph{International
  conference on intelligent computing}}. Springer, \bibinfo{pages}{878--887}.
\newblock


\bibitem[\protect\citeauthoryear{Han, Wang, and Mao}{Han
  et~al\mbox{.}}{2005b}]%
        {han2005borderline}
\bibfield{author}{\bibinfo{person}{Hui Han}, \bibinfo{person}{Wen-Yuan Wang},
  {and} \bibinfo{person}{Bing-Huan Mao}.} \bibinfo{year}{2005}\natexlab{b}.
\newblock \showarticletitle{Borderline-SMOTE: a new over-sampling method in
  imbalanced data sets learning}. In \bibinfo{booktitle}{\emph{International
  conference on intelligent computing}}. Springer, \bibinfo{pages}{878--887}.
\newblock


\bibitem[\protect\citeauthoryear{He, Bai, Garcia, and Li}{He
  et~al\mbox{.}}{2008a}]%
        {he2008adaptive}
\bibfield{author}{\bibinfo{person}{H He}, \bibinfo{person}{Y Bai},
  \bibinfo{person}{E Garcia}, {and} \bibinfo{person}{S~ADASYN Li}.}
  \bibinfo{year}{2008}\natexlab{a}.
\newblock \bibinfo{title}{Adaptive synthetic sampling approach for imbalanced
  learning. IEEE International Joint Conference on Neural Networks, 2008 (IEEE
  World Congress on Computational Intelligence)}.
\newblock
\newblock


\bibitem[\protect\citeauthoryear{He, Bai, Garcia, and Li}{He
  et~al\mbox{.}}{2008b}]%
        {adasyn}
\bibfield{author}{\bibinfo{person}{Haibo He}, \bibinfo{person}{Yang Bai},
  \bibinfo{person}{Edwardo~A Garcia}, {and} \bibinfo{person}{Shutao Li}.}
  \bibinfo{year}{2008}\natexlab{b}.
\newblock \showarticletitle{ADASYN: Adaptive synthetic sampling approach for
  imbalanced learning}. In \bibinfo{booktitle}{\emph{2008 IEEE International
  Joint Conference on Neural Networks (IEEE World Congress on Computational
  Intelligence)}}. IEEE, \bibinfo{pages}{1322--1328}.
\newblock


\bibitem[\protect\citeauthoryear{Kubat, Matwin, et~al\mbox{.}}{Kubat
  et~al\mbox{.}}{1997a}]%
        {kubat1997addressing}
\bibfield{author}{\bibinfo{person}{Miroslav Kubat}, \bibinfo{person}{Stan
  Matwin}, {et~al\mbox{.}}} \bibinfo{year}{1997}\natexlab{a}.
\newblock \showarticletitle{Addressing the curse of imbalanced training sets:
  one-sided selection}. In \bibinfo{booktitle}{\emph{Icml}},
  Vol.~\bibinfo{volume}{97}. Nashville, USA, \bibinfo{pages}{179--186}.
\newblock


\bibitem[\protect\citeauthoryear{Kubat, Matwin, et~al\mbox{.}}{Kubat
  et~al\mbox{.}}{1997b}]%
        {gmeanicml}
\bibfield{author}{\bibinfo{person}{Miroslav Kubat}, \bibinfo{person}{Stan
  Matwin}, {et~al\mbox{.}}} \bibinfo{year}{1997}\natexlab{b}.
\newblock \showarticletitle{Addressing the curse of imbalanced training sets:
  one-sided selection}. In \bibinfo{booktitle}{\emph{Icml}},
  Vol.~\bibinfo{volume}{97}. Nashville, USA, \bibinfo{pages}{179--186}.
\newblock


\bibitem[\protect\citeauthoryear{Miri~Rostami and Ahmadzadeh}{Miri~Rostami and
  Ahmadzadeh}{2018}]%
        {miri2018extracting}
\bibfield{author}{\bibinfo{person}{S Miri~Rostami} {and} \bibinfo{person}{M
  Ahmadzadeh}.} \bibinfo{year}{2018}\natexlab{}.
\newblock \showarticletitle{Extracting predictor variables to construct breast
  cancer survivability model with class imbalance problem}.
\newblock \bibinfo{journal}{\emph{Journal of AI and Data Mining}}
  \bibinfo{volume}{6}, \bibinfo{number}{2} (\bibinfo{year}{2018}),
  \bibinfo{pages}{263--276}.
\newblock


\bibitem[\protect\citeauthoryear{Pedregosa, Varoquaux, Gramfort, Michel,
  Thirion, Grisel, Blondel, Prettenhofer, Weiss, Dubourg,
  et~al\mbox{.}}{Pedregosa et~al\mbox{.}}{2011}]%
        {pedregosa2011scikit}
\bibfield{author}{\bibinfo{person}{Fabian Pedregosa}, \bibinfo{person}{Ga{\"e}l
  Varoquaux}, \bibinfo{person}{Alexandre Gramfort}, \bibinfo{person}{Vincent
  Michel}, \bibinfo{person}{Bertrand Thirion}, \bibinfo{person}{Olivier
  Grisel}, \bibinfo{person}{Mathieu Blondel}, \bibinfo{person}{Peter
  Prettenhofer}, \bibinfo{person}{Ron Weiss}, \bibinfo{person}{Vincent
  Dubourg}, {et~al\mbox{.}}} \bibinfo{year}{2011}\natexlab{}.
\newblock \showarticletitle{Scikit-learn: Machine learning in Python}.
\newblock \bibinfo{journal}{\emph{Journal of machine learning research}}
  \bibinfo{volume}{12}, \bibinfo{number}{Oct} (\bibinfo{year}{2011}),
  \bibinfo{pages}{2825--2830}.
\newblock


\bibitem[\protect\citeauthoryear{Sharma, Bellinger, Krawczyk, Zaiane, and
  Japkowicz}{Sharma et~al\mbox{.}}{2018}]%
        {swim}
\bibfield{author}{\bibinfo{person}{Shiven Sharma}, \bibinfo{person}{Colin
  Bellinger}, \bibinfo{person}{Bartosz Krawczyk}, \bibinfo{person}{Osmar
  Zaiane}, {and} \bibinfo{person}{Nathalie Japkowicz}.}
  \bibinfo{year}{2018}\natexlab{}.
\newblock \showarticletitle{Synthetic oversampling with the majority class: A
  new perspective on handling extreme imbalance}. In
  \bibinfo{booktitle}{\emph{2018 IEEE International Conference on Data Mining
  (ICDM)}}. IEEE, \bibinfo{pages}{447--456}.
\newblock


\bibitem[\protect\citeauthoryear{Summers and Dinneen}{Summers and
  Dinneen}{2018}]%
        {mixed_eg}
\bibfield{author}{\bibinfo{person}{Cecilia Summers} {and}
  \bibinfo{person}{Michael~J. Dinneen}.} \bibinfo{year}{2018}\natexlab{}.
\newblock \showarticletitle{Improved Mixed-Example Data Augmentation}.
\newblock \bibinfo{journal}{\emph{2019 IEEE Winter Conference on Applications
  of Computer Vision (WACV)}} (\bibinfo{year}{2018}),
  \bibinfo{pages}{1262--1270}.
\newblock


\bibitem[\protect\citeauthoryear{Summers and Dinneen}{Summers and
  Dinneen}{2019}]%
        {summers2019improved}
\bibfield{author}{\bibinfo{person}{Cecilia Summers} {and}
  \bibinfo{person}{Michael~J Dinneen}.} \bibinfo{year}{2019}\natexlab{}.
\newblock \showarticletitle{Improved mixed-example data augmentation}. In
  \bibinfo{booktitle}{\emph{2019 IEEE Winter Conference on Applications of
  Computer Vision (WACV)}}. IEEE, \bibinfo{pages}{1262--1270}.
\newblock


\bibitem[\protect\citeauthoryear{Tanaka and Aranha}{Tanaka and Aranha}{2019}]%
        {gan1}
\bibfield{author}{\bibinfo{person}{Fabio Henrique Kiyoiti dos~Santos Tanaka}
  {and} \bibinfo{person}{Claus Aranha}.} \bibinfo{year}{2019}\natexlab{}.
\newblock \showarticletitle{Data Augmentation Using GANs}.
\newblock \bibinfo{journal}{\emph{arXiv preprint arXiv:1904.09135}}
  (\bibinfo{year}{2019}).
\newblock


\bibitem[\protect\citeauthoryear{Tokozume, Ushiku, and Harada}{Tokozume
  et~al\mbox{.}}{2017}]%
        {bcplus}
\bibfield{author}{\bibinfo{person}{Yuji Tokozume}, \bibinfo{person}{Yoshitaka
  Ushiku}, {and} \bibinfo{person}{Tatsuya Harada}.}
  \bibinfo{year}{2017}\natexlab{}.
\newblock \showarticletitle{Between-Class Learning for Image Classification}.
\newblock \bibinfo{journal}{\emph{2018 IEEE/CVF Conference on Computer Vision
  and Pattern Recognition}} (\bibinfo{year}{2017}),
  \bibinfo{pages}{5486--5494}.
\newblock


\bibitem[\protect\citeauthoryear{Tokozume, Ushiku, and Harada}{Tokozume
  et~al\mbox{.}}{2018}]%
        {bc+}
\bibfield{author}{\bibinfo{person}{Yuji Tokozume}, \bibinfo{person}{Yoshitaka
  Ushiku}, {and} \bibinfo{person}{Tatsuya Harada}.}
  \bibinfo{year}{2018}\natexlab{}.
\newblock \showarticletitle{Between-class learning for image classification}.
  In \bibinfo{booktitle}{\emph{Proceedings of the IEEE Conference on Computer
  Vision and Pattern Recognition}}. \bibinfo{pages}{5486--5494}.
\newblock


\bibitem[\protect\citeauthoryear{Wang, Gao, Shi, and Wang}{Wang
  et~al\mbox{.}}{2016}]%
        {wang2016fast}
\bibfield{author}{\bibinfo{person}{Huihui Wang}, \bibinfo{person}{Yang Gao},
  \bibinfo{person}{Yinghuan Shi}, {and} \bibinfo{person}{Hao Wang}.}
  \bibinfo{year}{2016}\natexlab{}.
\newblock \showarticletitle{A fast distributed classification algorithm for
  large-scale imbalanced data}. In \bibinfo{booktitle}{\emph{2016 IEEE 16th
  International Conference on Data Mining (ICDM)}}. IEEE,
  \bibinfo{pages}{1251--1256}.
\newblock


\bibitem[\protect\citeauthoryear{Wang, Minku, and Yao}{Wang
  et~al\mbox{.}}{2014}]%
        {wang2014resampling}
\bibfield{author}{\bibinfo{person}{Shuo Wang}, \bibinfo{person}{Leandro~L
  Minku}, {and} \bibinfo{person}{Xin Yao}.} \bibinfo{year}{2014}\natexlab{}.
\newblock \showarticletitle{Resampling-based ensemble methods for online class
  imbalance learning}.
\newblock \bibinfo{journal}{\emph{IEEE Transactions on Knowledge and Data
  Engineering}} \bibinfo{volume}{27}, \bibinfo{number}{5}
  (\bibinfo{year}{2014}), \bibinfo{pages}{1356--1368}.
\newblock


\bibitem[\protect\citeauthoryear{Zhang, Cisse, Dauphin, and Lopez-Paz}{Zhang
  et~al\mbox{.}}{2017}]%
        {mixup}
\bibfield{author}{\bibinfo{person}{Hongyi Zhang}, \bibinfo{person}{Moustapha
  Cisse}, \bibinfo{person}{Yann~N Dauphin}, {and} \bibinfo{person}{David
  Lopez-Paz}.} \bibinfo{year}{2017}\natexlab{}.
\newblock \showarticletitle{mixup: Beyond empirical risk minimization}.
\newblock \bibinfo{journal}{\emph{arXiv preprint arXiv:1710.09412}}
  (\bibinfo{year}{2017}).
\newblock


\end{thebibliography}


\begin{thebibliography}{00}

%%% the Web address.



\ifx \showCODEN    \undefined \def \showCODEN     #1{\unskip}     \fi
\ifx \showDOI      \undefined \def \showDOI       #1{#1}\fi
\ifx \showISBNx    \undefined \def \showISBNx     #1{\unskip}     \fi
\ifx \showISBNxiii \undefined \def \showISBNxiii  #1{\unskip}     \fi
\ifx \showISSN     \undefined \def \showISSN      #1{\unskip}     \fi
\ifx \showLCCN     \undefined \def \showLCCN      #1{\unskip}     \fi
\ifx \shownote     \undefined \def \shownote      #1{#1}          \fi
\ifx \showarticletitle \undefined \def \showarticletitle #1{#1}   \fi
\ifx \showURL      \undefined \def \showURL       {\relax}        \fi
% invisible to TeX
\providecommand\bibfield[2]{#2}
\providecommand\bibinfo[2]{#2}
\providecommand\natexlab[1]{#1}
\providecommand\showeprint[2][]{arXiv:#2}


\bibitem{abdi2015combat}
\bibfield{author}{\bibinfo{person}{Lida Abdi} {and} \bibinfo{person}{Sattar
  Hashemi}.} \bibinfo{year}{2015}\natexlab{}.
\newblock \showarticletitle{To combat multi-class imbalanced problems by means
  of over-sampling techniques}.
\newblock \bibinfo{journal}{\emph{IEEE transactions on Knowledge and Data
  Engineering}} \bibinfo{volume}{28}, \bibinfo{number}{1}
  (\bibinfo{year}{2015}), \bibinfo{pages}{238--251}.
\newblock


\bibitem%
        {al2005feature}
\bibfield{author}{\bibinfo{person}{Ali Al-Shahib}, \bibinfo{person}{Rainer
  Breitling}, {and} \bibinfo{person}{David Gilbert}.}
  \bibinfo{year}{2005}\natexlab{}.
\newblock \showarticletitle{Feature selection and the class imbalance problem
  in predicting protein function from sequence}.
\newblock \bibinfo{journal}{\emph{Applied Bioinformatics}} \bibinfo{volume}{4},
  \bibinfo{number}{3} (\bibinfo{year}{2005}), \bibinfo{pages}{195--203}.
\newblock


\bibitem%
        {bayes_sign_test}
\bibfield{author}{\bibinfo{person}{Alessio Benavoli}, \bibinfo{person}{Giorgio
  Corani}, \bibinfo{person}{Francesca Mangili}, \bibinfo{person}{Marco
  Zaffalon}, {and} \bibinfo{person}{Fabrizio Ruggeri}.}
  \bibinfo{year}{2014}\natexlab{}.
\newblock \showarticletitle{A Bayesian Wilcoxon signed-rank test based on the
  Dirichlet process.}. In \bibinfo{booktitle}{\emph{ICML}}
  \emph{(\bibinfo{series}{JMLR Workshop and Conference Proceedings})},
  Vol.~\bibinfo{volume}{32}. \bibinfo{publisher}{JMLR.org},
  \bibinfo{pages}{1026--1034}.
\newblock
\urldef\tempurl%
\url{http://dblp.uni-trier.de/db/conf/icml/icml2014.html#BenavoliCMZR14}
\showURL{%
\tempurl}


\bibitem%
        {bennin2017mahakil}
\bibfield{author}{\bibinfo{person}{Kwabena~Ebo Bennin}, \bibinfo{person}{Jacky
  Keung}, \bibinfo{person}{Passakorn Phannachitta}, \bibinfo{person}{Akito
  Monden}, {and} \bibinfo{person}{Solomon Mensah}.}
  \bibinfo{year}{2017}\natexlab{}.
\newblock \showarticletitle{Mahakil: Diversity based oversampling approach to
  alleviate the class imbalance issue in software defect prediction}.
\newblock \bibinfo{journal}{\emph{IEEE Transactions on Software Engineering}}
  \bibinfo{volume}{44}, \bibinfo{number}{6} (\bibinfo{year}{2017}),
  \bibinfo{pages}{534--550}.
\newblock


% \bibitem%
%         {burez2009handling}
% \bibfield{author}{\bibinfo{person}{Jonathan Burez} {and} \bibinfo{person}{Dirk
%   Van~den Poel}.} \bibinfo{year}{2009}\natexlab{}.
% \newblock \showarticletitle{Handling class imbalance in customer churn
%   prediction}.
% \newblock \bibinfo{journal}{\emph{Expert Systems with Applications}}
%   \bibinfo{volume}{36}, \bibinfo{number}{3} (\bibinfo{year}{2009}),
%   \bibinfo{pages}{4626--4636}.
% \newblock


\bibitem%
        {chawla2002smote}
\bibfield{author}{\bibinfo{person}{Nitesh~V Chawla}, \bibinfo{person}{Kevin~W
  Bowyer}, \bibinfo{person}{Lawrence~O Hall}, {and} \bibinfo{person}{W~Philip
  Kegelmeyer}.} \bibinfo{year}{2002}\natexlab{}.
\newblock \showarticletitle{SMOTE: synthetic minority over-sampling technique}.
\newblock \bibinfo{journal}{\emph{Journal of artificial intelligence research}}
   \bibinfo{volume}{16} (\bibinfo{year}{2002}), \bibinfo{pages}{321--357}.
\newblock


\bibitem%
        {chawla2003smoteboost}
\bibfield{author}{\bibinfo{person}{Nitesh~V Chawla},
  \bibinfo{person}{Aleksandar Lazarevic}, \bibinfo{person}{Lawrence~O Hall},
  {and} \bibinfo{person}{Kevin~W Bowyer}.} \bibinfo{year}{2003}\natexlab{}.
\newblock \showarticletitle{SMOTEBoost: Improving prediction of the minority
  class in boosting}. In \bibinfo{booktitle}{\emph{European conference on
  principles of data mining and knowledge discovery}}. Springer,
  \bibinfo{pages}{107--119}.
\newblock


% \bibitem%
%         {github_code}
% \bibfield{author}{\bibinfo{person}{Anonymous~MixBoost Code}.}
%   \bibinfo{year}{2019}\natexlab{}.
% \newblock \bibinfo{title}{MixBoost-Code}.
% \newblock
%   \bibinfo{howpublished}{\url{https://github.com/user1332/MIXBOOST-code}}.
% \newblock


\bibitem%
        {uci_archive}
\bibfield{author}{\bibinfo{person}{Dheeru Dua} {and} \bibinfo{person}{Casey
  Graff}.} \bibinfo{year}{2017}\natexlab{}.
\newblock \bibinfo{title}{{UCI} Machine Learning Repository}.
\newblock
\newblock
\urldef\tempurl%
\url{http://archive.ics.uci.edu/ml}
\showURL{%
\tempurl}


\bibitem%
        {elkan2001foundations}
\bibfield{author}{\bibinfo{person}{Charles Elkan}.}
  \bibinfo{year}{2001}\natexlab{}.
\newblock \showarticletitle{The foundations of cost-sensitive learning}. In
  \bibinfo{booktitle}{\emph{International joint conference on artificial
  intelligence}}, Vol.~\bibinfo{volume}{17}. Lawrence Erlbaum Associates Ltd,
  \bibinfo{pages}{973--978}.
\newblock


\bibitem%
        {fernandez2018smote}
\bibfield{author}{\bibinfo{person}{Alberto Fern{\'a}ndez},
  \bibinfo{person}{Salvador Garcia}, \bibinfo{person}{Francisco Herrera}, {and}
  \bibinfo{person}{Nitesh~V Chawla}.} \bibinfo{year}{2018}\natexlab{}.
\newblock \showarticletitle{SMOTE for learning from imbalanced data: progress
  and challenges, marking the 15-year anniversary}.
\newblock \bibinfo{journal}{\emph{Journal of artificial intelligence research}}
   \bibinfo{volume}{61} (\bibinfo{year}{2018}), \bibinfo{pages}{863--905}.
\newblock


\bibitem%
        {galar2011review}
\bibfield{author}{\bibinfo{person}{Mikel Galar}, \bibinfo{person}{Alberto
  Fernandez}, \bibinfo{person}{Edurne Barrenechea}, \bibinfo{person}{Humberto
  Bustince}, {and} \bibinfo{person}{Francisco Herrera}.}
  \bibinfo{year}{2011}\natexlab{}.
\newblock \showarticletitle{A review on ensembles for the class imbalance
  problem: bagging-, boosting-, and hybrid-based approaches}.
\newblock \bibinfo{journal}{\emph{IEEE Transactions on Systems, Man, and
  Cybernetics, Part C (Applications and Reviews)}} \bibinfo{volume}{42},
  \bibinfo{number}{4} (\bibinfo{year}{2011}), \bibinfo{pages}{463--484}.
\newblock


\bibitem%
        {garcia2009evolutionary}
\bibfield{author}{\bibinfo{person}{Salvador Garc{\'\i}a} {and}
  \bibinfo{person}{Francisco Herrera}.} \bibinfo{year}{2009}\natexlab{}.
\newblock \showarticletitle{Evolutionary undersampling for classification with
  imbalanced datasets: Proposals and taxonomy}.
\newblock \bibinfo{journal}{\emph{Evolutionary computation}}
  \bibinfo{volume}{17}, \bibinfo{number}{3} (\bibinfo{year}{2009}),
  \bibinfo{pages}{275--306}.
\newblock


\bibitem%
        {borderlinesmote}
\bibfield{author}{\bibinfo{person}{Hui Han}, \bibinfo{person}{Wen-Yuan Wang},
  {and} \bibinfo{person}{Bing-Huan Mao}.} \bibinfo{year}{2005}\natexlab{a}.
\newblock \showarticletitle{Borderline-SMOTE: a new over-sampling method in
  imbalanced data sets learning}. In \bibinfo{booktitle}{\emph{International
  conference on intelligent computing}}. Springer, \bibinfo{pages}{878--887}.
\newblock


\bibitem%
        {he2008adaptive}
\bibfield{author}{\bibinfo{person}{H He}, \bibinfo{person}{Y Bai},
  \bibinfo{person}{E Garcia}, {and} \bibinfo{person}{S~ADASYN Li}.}
  \bibinfo{year}{2008}\natexlab{a}.
\newblock \bibinfo{title}{Adaptive synthetic sampling approach for imbalanced
  learning. IEEE International Joint Conference on Neural Networks, 2008 (IEEE
  World Congress on Computational Intelligence)}.
\newblock
\newblock


\bibitem%
        {adasyn}
\bibfield{author}{\bibinfo{person}{Haibo He}, \bibinfo{person}{Yang Bai},
  \bibinfo{person}{Edwardo~A Garcia}, {and} \bibinfo{person}{Shutao Li}.}
  \bibinfo{year}{2008}\natexlab{b}.
\newblock \showarticletitle{ADASYN: Adaptive synthetic sampling approach for
  imbalanced learning}. In \bibinfo{booktitle}{\emph{2008 IEEE International
  Joint Conference on Neural Networks (IEEE World Congress on Computational
  Intelligence)}}. IEEE, \bibinfo{pages}{1322--1328}.
\newblock


\bibitem%
        {kubat1997addressing}
\bibfield{author}{\bibinfo{person}{Miroslav Kubat}, \bibinfo{person}{Stan
  Matwin}, {et~al\mbox{.}}} \bibinfo{year}{1997}\natexlab{a}.
\newblock \showarticletitle{Addressing the curse of imbalanced training sets:
  one-sided selection}. In \bibinfo{booktitle}{\emph{Icml}},
  Vol.~\bibinfo{volume}{97}. Nashville, USA, \bibinfo{pages}{179--186}.
\newblock


\bibitem%
        {miri2018extracting}
\bibfield{author}{\bibinfo{person}{S Miri~Rostami} {and} \bibinfo{person}{M
  Ahmadzadeh}.} \bibinfo{year}{2018}\natexlab{}.
\newblock \showarticletitle{Extracting predictor variables to construct breast
  cancer survivability model with class imbalance problem}.
\newblock \bibinfo{journal}{\emph{Journal of AI and Data Mining}}
  \bibinfo{volume}{6}, \bibinfo{number}{2} (\bibinfo{year}{2018}),
  \bibinfo{pages}{263--276}.
\newblock


\bibitem%
        {pedregosa2011scikit}
\bibfield{author}{\bibinfo{person}{Fabian Pedregosa}, \bibinfo{person}{Ga{\"e}l
  Varoquaux}, \bibinfo{person}{Alexandre Gramfort}, \bibinfo{person}{Vincent
  Michel}, \bibinfo{person}{Bertrand Thirion}, \bibinfo{person}{Olivier
  Grisel}, \bibinfo{person}{Mathieu Blondel}, \bibinfo{person}{Peter
  Prettenhofer}, \bibinfo{person}{Ron Weiss}, \bibinfo{person}{Vincent
  Dubourg}, {et~al\mbox{.}}} \bibinfo{year}{2011}\natexlab{}.
\newblock \showarticletitle{Scikit-learn: Machine learning in Python}.
\newblock \bibinfo{journal}{\emph{Journal of machine learning research}}
  \bibinfo{volume}{12}, \bibinfo{number}{Oct} (\bibinfo{year}{2011}),
  \bibinfo{pages}{2825--2830}.
\newblock


\bibitem%
        {swim}
\bibfield{author}{\bibinfo{person}{Shiven Sharma}, \bibinfo{person}{Colin
  Bellinger}, \bibinfo{person}{Bartosz Krawczyk}, \bibinfo{person}{Osmar
  Zaiane}, {and} \bibinfo{person}{Nathalie Japkowicz}.}
  \bibinfo{year}{2018}\natexlab{}.
\newblock \showarticletitle{Synthetic oversampling with the majority class: A
  new perspective on handling extreme imbalance}. In
  \bibinfo{booktitle}{\emph{2018 IEEE International Conference on Data Mining
  (ICDM)}}. IEEE, \bibinfo{pages}{447--456}.
\newblock


% mixed_eg
\bibitem%
        {summers2019improved}
\bibfield{author}{\bibinfo{person}{Cecilia Summers} {and}
  \bibinfo{person}{Michael~J Dinneen}.} \bibinfo{year}{2019}\natexlab{}.
\newblock \showarticletitle{Improved mixed-example data augmentation}. In
  \bibinfo{booktitle}{\emph{2019 IEEE Winter Conference on Applications of
  Computer Vision (WACV)}}. IEEE, \bibinfo{pages}{1262--1270}.
\newblock


\bibitem%
        {gan1}
\bibfield{author}{\bibinfo{person}{Fabio Henrique Kiyoiti dos~Santos Tanaka}
  {and} \bibinfo{person}{Claus Aranha}.} \bibinfo{year}{2019}\natexlab{}.
\newblock \showarticletitle{Data Augmentation Using GANs}.
\newblock \bibinfo{journal}{\emph{arXiv preprint arXiv:1904.09135}}
  (\bibinfo{year}{2019}).
\newblock


\bibitem%
        {bcplus}
\bibfield{author}{\bibinfo{person}{Yuji Tokozume}, \bibinfo{person}{Yoshitaka
  Ushiku}, {and} \bibinfo{person}{Tatsuya Harada}.}
  \bibinfo{year}{2017}\natexlab{}.
\newblock \showarticletitle{Between-Class Learning for Image Classification}.
\newblock \bibinfo{journal}{\emph{2018 IEEE/CVF Conference on Computer Vision
  and Pattern Recognition}} (\bibinfo{year}{2017}),
  \bibinfo{pages}{5486--5494}.
\newblock


\bibitem%
        {wang2016fast}
\bibfield{author}{\bibinfo{person}{Huihui Wang}, \bibinfo{person}{Yang Gao},
  \bibinfo{person}{Yinghuan Shi}, {and} \bibinfo{person}{Hao Wang}.}
  \bibinfo{year}{2016}\natexlab{}.
\newblock \showarticletitle{A fast distributed classification algorithm for
  large-scale imbalanced data}. In \bibinfo{booktitle}{\emph{2016 IEEE 16th
  International Conference on Data Mining (ICDM)}}. IEEE,
  \bibinfo{pages}{1251--1256}.
\newblock


\bibitem%
        {wang2014resampling}
\bibfield{author}{\bibinfo{person}{Shuo Wang}, \bibinfo{person}{Leandro~L
  Minku}, {and} \bibinfo{person}{Xin Yao}.} \bibinfo{year}{2014}\natexlab{}.
\newblock \showarticletitle{Resampling-based ensemble methods for online class
  imbalance learning}.
\newblock \bibinfo{journal}{\emph{IEEE Transactions on Knowledge and Data
  Engineering}} \bibinfo{volume}{27}, \bibinfo{number}{5}
  (\bibinfo{year}{2014}), \bibinfo{pages}{1356--1368}.
\newblock


\bibitem%
        {mixup}
\bibfield{author}{\bibinfo{person}{Hongyi Zhang}, \bibinfo{person}{Moustapha
  Cisse}, \bibinfo{person}{Yann~N Dauphin}, {and} \bibinfo{person}{David
  Lopez-Paz}.} \bibinfo{year}{2017}\natexlab{}.
\newblock \showarticletitle{mixup: Beyond empirical risk minimization}.
\newblock \bibinfo{journal}{\emph{arXiv preprint arXiv:1710.09412}}
  (\bibinfo{year}{2017}).
\newblock

\bibitem%
        {roc_auc_ref}
\bibfield{author}{\bibinfo{person}{Andrew~P. Bradley}.}
  \bibinfo{year}{1997}\natexlab{}.
\newblock \showarticletitle{The Use of the Area under the ROC Curve in the
  Evaluation of Machine Learning Algorithms}.
\newblock \bibinfo{journal}{\emph{Pattern Recogn.}} \bibinfo{volume}{30},
  \bibinfo{number}{7} (\bibinfo{date}{July} \bibinfo{year}{1997}),
  \bibinfo{pages}{1145–1159}.
\newblock
\showISSN{0031-3203}
\urldef\tempurl%
\url{https://doi.org/10.1016/S0031-3203(96)00142-2}
\showDOI{\tempurl}



\bibitem%
        {sharma2018learning}
\bibfield{author}{\bibinfo{person}{Shiven Sharma}, \bibinfo{person}{Anil
  Somayaji}, {and} \bibinfo{person}{Nathalie Japkowicz}.}
  \bibinfo{year}{2018}\natexlab{b}.
\newblock \showarticletitle{Learning over subconcepts: Strategies for 1-class
  classification}.
\newblock \bibinfo{journal}{\emph{Computational Intelligence}}
  \bibinfo{volume}{34}, \bibinfo{number}{2} (\bibinfo{year}{2018}),
  \bibinfo{pages}{440--467}.
\newblock


\bibitem%
        {suhandy2017use}
\bibfield{author}{\bibinfo{person}{Diding Suhandy} {and}
  \bibinfo{person}{Meinilwita Yulia}.} \bibinfo{year}{2017}\natexlab{}.
\newblock \showarticletitle{The use of partial least square regression and
  spectral data in UV-visible region for quantification of adulteration in
  Indonesian palm civet coffee}.
\newblock \bibinfo{journal}{\emph{International journal of food science}}
  \bibinfo{volume}{2017} (\bibinfo{year}{2017}).
\newblock


\bibitem%
        {wei2013effective}
\bibfield{author}{\bibinfo{person}{Wei Wei}, \bibinfo{person}{Jinjiu Li},
  \bibinfo{person}{Longbing Cao}, \bibinfo{person}{Yuming Ou}, {and}
  \bibinfo{person}{Jiahang Chen}.} \bibinfo{year}{2013}\natexlab{}.
\newblock \showarticletitle{Effective detection of sophisticated online banking
  fraud on extremely imbalanced data}.
\newblock \bibinfo{journal}{\emph{World Wide Web}} \bibinfo{volume}{16},
  \bibinfo{number}{4} (\bibinfo{year}{2013}), \bibinfo{pages}{449--475}.
\newblock

\end{thebibliography}
% \input{output.bbl}

% \newcommand{\showURL}[1]{\unskip}
% \newcommand{\showarticletitle}[1]{\unskip}
% \newcommand{\showCODEN}[1]{\unskip}
% \newcommand{\showDOI}[1]{\unskip}
\newcommand{\showISBNx}[1]{\unskip}
\newcommand{\showLCCN}[1]{\unskip}

\end{document}